\documentclass{article} 
\usepackage{iclr2027_conference,times}

\ifdefined\XeTeXrevision\else\ifdefined\directlua\else
  \usepackage[utf8]{inputenc}
  \usepackage[T1]{fontenc}
\fi\fi
\IfFileExists{microtype.sty}{\usepackage{microtype}}{}

\usepackage{booktabs}
\usepackage{graphicx}
\usepackage{amsmath}
\usepackage{amssymb}
\usepackage{amsthm}
\usepackage{xcolor}
\usepackage{colortbl}
\usepackage{xspace}
\usepackage{natbib}          

\IfFileExists{tcolorbox.sty}{%
  \usepackage{tcolorbox}%
  \newtcolorbox{theorybox}{%
    colback=gray!6, colframe=black!70, boxrule=0.6pt, arc=1.5mm,
    left=7pt, right=7pt, top=6pt, bottom=6pt}%
}{%
  \usepackage{framed}%
  \newenvironment{theorybox}{\begin{framed}}{\end{framed}}%
}

\usepackage[colorlinks=true,
            linkcolor=blue!55!black,
            citecolor=green!40!black,
            urlcolor=blue!55!black]{hyperref}
\usepackage[capitalize,noabbrev]{cleveref}

\theoremstyle{plain}

\theoremstyle{definition}
\newtheorem{definition}{Definition}

\theoremstyle{remark}

\newif\ifFrozenResults
\FrozenResultstrue   

\newcommand{\method}{TopoSHAP\xspace}

\newcommand{\expclaim}[1]{\noindent\textbf{#1}}

\newcommand{\eg}{e.g.,\xspace}

\newcommand{\rk}{\operatorname{rk}}

\newcommand{\ResNumNeighborhoods}{11}         

\newcommand{\ResExactnessMaxErr}{$2.254\times 10^{-14}$} 

\usepackage{amsmath,amsfonts,bm}

\def\eqref#1{equation~\ref{#1}}

\def\1{\bm{1}}

\def\rk{{\textnormal{k}}}

\DeclareMathAlphabet{\mathsfit}{\encodingdefault}{\sfdefault}{m}{sl}
\SetMathAlphabet{\mathsfit}{bold}{\encodingdefault}{\sfdefault}{bx}{n}

\usepackage{hyperref}
\usepackage{url}
\usepackage{enumitem}

\usepackage{algorithm}
\usepackage{algpseudocode}
\usepackage{amsmath}
\usepackage{tabularx}

\title{TopoSHAP: Shapley for 
  Explainable \\ and Efficient Topological Deep Learning}

\author{Mathilde Papillon \\
Department of Physics\\
University of California, Santa Barbara\\
\texttt{papillon@ucsb.edu}
\And
Nina Miolane \\
Department of Electrical and Computer Engineering\\
University of California, Santa Barbara\\
}

\iclrfinalcopy
\begin{document}

\maketitle
\lhead{ } 
\begin{abstract}
Topological deep learning (TDL) extends learning on graphs, such as molecules or social networks, to higher-order structures, such as rings in a molecule or groups in a social network. Explaining which of these structures drive a model's predictions, and which architectural components matter most for its performance, remains challenging. We introduce TopoSHAP, an efficient and general explanation framework that addresses both questions across topological domains and model architectures. TopoSHAP-cell assigns Shapley values to topological data cells and extracts compact explanations. These explanations match or outperform graph-based counterparts on GraphXAI benchmark data while running up to 112$\times$ faster. They also reveal advantageous topological feature design, leading to measured improvements in explanation recovery and predictive accuracy. On the architecture side, TopoSHAP-neighbor explains how neighborhoods contribute to model performance, and turns these evaluations into a backward-elimination search that identifies smaller, competitive architectures early in training. This is up to $11.3\times$ cheaper than traditional TDL training sweeps.\end{abstract}

\section{Introduction}
\label{sec:introduction}

Graph neural networks (GNNs) learn from relational data, such as molecules and social networks \citep{kipf2017semisupervised}. They achieve this by explicitly modeling pairwise interactions between the nodes (atoms, people) of the data. However, real-world systems often exhibit interactions that are inherently multiway. For instance, atoms can form rings which have critical functional roles; people can form groups whose nature goes beyond pairwise relationships. These higher-order interactions are not explicitly represented in standard GNNs. Topological neural networks (TNNs) address this limitation by explicitly modeling higher-order structure \citep{hajij2023tdl,papillon2023architectures}. In particular, topological deep learning (TDL) uses \emph{combinatorial complexes} to organize interactions into \emph{cells} of different \emph{ranks}: atoms are rank-0 cells, bonds are rank-1 cells, and rings are rank-2 cells. At each layer, TNNs update each cell feature based on the features of its neighboring cells. On standardized topological benchmarks, TNNs often match or exceed GNN accuracy \citep{telyatnikov2024topobench}.

As TDL moves towards domain-specific applications like chemistry and materials science \citep{papamarkou2024position}, practitioners face two black box challenges. First, they need to understand which structures drive these predictions. For a molecular model, for example, one may be compelled to reveal the chemical features and molecular structures the model deems relevant to the target property. Second, TDL practitioners do not know \textit{a priori} what topological relations, or \textit{neighborhoods}, matter for their task. Current TDL pipelines rely on an expensive hyperparameter search to identify the best architecture \citep{papillon2025topotune,hopse2025}. This is not only expensive but also fails to provide practical insight as to why a specific architecture worked better. 

Machine learning \emph{explainers} are a promising avenue to opening these black boxes. They investigate model behavior by attributing predictions to parts of the input \citep{doshivelez2017rigorous,ribeiro2016trust,lundberg2017shap}. However, existing post-hoc explainers are limited to graph \citep{sinha2025forge,ying2019gnnexplainer,luo2020parameterized,yuan2021subgraphx,muschalik2024graphshapiq} and hypergraph \citep{maleki2023hyperex,su2024shypx} neural networks. We propose the first explainer for general topological domains, encompassing both prediction and performance explanation. 

\paragraph{Contributions.}
We introduce \method, a unified game-theoretic framework for explaining TDL models, which attributes both predictions and performance to the cells and neighborhoods of a combinatorial complex. \method is available at \href{https://github.com/mathildepapillon/toposhap}{github.com/mathildepapillon/toposhap}.
\begin{itemize}[leftmargin=*]
\item \textbf{We explain predictions across ranks.} We formulate a cooperative game in which cells of every rank are players, producing per-cell Shapley attributions and compact explanations. Across both outputs, we match or outperform graph explanation methods on GraphXAI benchmarks. We also propose the first explanations of topological benchmark datasets going beyond hypergraphs. We call this component \method-cell.
\item \textbf{We use cell attributions to improve topological features.} Rank-specific attributions show that topological features, typically built by aggregating node features, retain information about lower-rank cells. Using this finding, we design new topological features and match or improve explanation recovery on GraphXAI benchmarks.
\item \textbf{We explain performance and use it to select architectures.} We formulate a second game in which a model's neighborhood-specific modules are players and test performance is the payoff. We propose an inexpensive backward-elimination procedure that selects neighborhood-specific modules for full training. This reduces training compute by up to an order of magnitude relative to standard TDL training sweeps. We call this component \method-neighbor.
\end{itemize}

\begin{figure}[t]
  \centering
  \includegraphics[width=\linewidth]{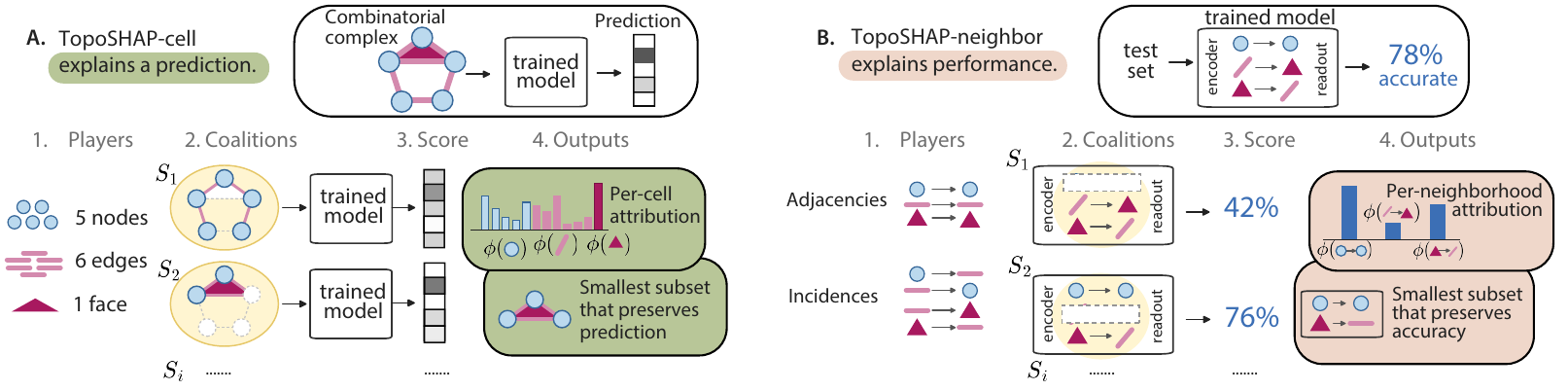}
  \caption{\textbf{\method explains a Topological Neural Network's predictions and performance.}
  \textbf{A.}~\method-cell considers the cells of the input combinatorial complex as players (1). Given certain coalitions $S_i$ of those players (2), it evaluates the model's predictions (3). Averaging marginal contributions over coalitions outputs one Shapley value per cell and a sub-complex best preserving the prediction (4). \textbf{B.}~\method-neighbor manipulates the neighborhoods that contribute to feature updates in the TNN (1). It briefly trains a model with all neighborhoods, then masks coalitions of them (2) to measure neighborhood impact on test performance (3). Averaging marginal contributions over coalitions outputs one attribution value per neighborhood and a subset of that preserves model performance (4).}
  \label{fig:overview}
\end{figure}

\section{Related Work}
\label{sec:related}

\method builds on three lines of work: (i) topological deep learning, the
subfield proposing the TNNs we explain; (ii) Shapley-value attribution in
machine learning, the game-theoretic background we rely on; and (iii) GNN
explainers, the methods we generalize. We review each below.

\paragraph{Topological deep learning (TDL).}
TDL proposes neural networks that learn from higher-order topological domains generalizing graphs, such as: simplicial complexes \citep{ebli2020simplicial}, cellular complexes \citep{bodnar2021cwn}, hypergraphs \citep{feng2019Hypergraph} and combinatorial complexes
\citep{hajij2023tdl}.
TopoTune \citep{papillon2025topotune} defines the architectural design space we study. Specifically, we consider Generalized Combinatorial Complex Neural Networks (GCCN) as they are general TNNs that are conveniently specified by the set of neighborhoods they each admit. We also consider the non-message passing version of GCCNs, introduced in HOPSE \citep{hopse2025}, which leverage graph embeddings. Our framework treats HOPSE and message-passing TNNs identically because it only requires forward evaluations. Regarding implementation, TopoBench \citep{telyatnikov2024topobench} standardizes liftings, models, and training across topological domains. We host our contribution and experiments there. 

\paragraph{Shapley-value attribution in machine learning.} The Shapley value credits each player in a cooperative game with its average marginal contribution across coalitions, and is the unique such credit assignment satisfying a standard set of fairness axioms \citep{shapley1953value}.
SHAP brought this machinery to machine learning with input features as players \citep{lundberg2017shap}, and interaction indices extend credit to coalitions \citep{grabisch1999axiomatic}. In \method-cell, our players are the cells (node, edge, hyperedge, face, and so on) that make up the topological domain. Because exact values cost exponential time in the number of players,  larger complexes use unbiased permutation sampling \citep{castro2009polynomial,strumbelj2010efficient}. Shapley values are unsound for selecting players when two players can do the same job \citep{fryer2021shapley}. Our framework therefore uses the game only to order candidates, and compares the shortlisted candidates by validation of fully trained models (\cref{sec:framework-game2}).

\paragraph{Explaining GNNs and Hypergraph Neural Networks.}
Existing GNN explainers broadly fall into two families. The first family of explainers learns continuous masks over the graph's nodes and edges. GNNExplainer learns a new mask for each instance, requiring a new optimization run per prediction \citep{ying2019gnnexplainer}. PGExplainer trains a single mask-generating model once over the dataset, which then generalizes to new instances without further optimization \citep{luo2020parameterized}.
Second, search-based explainers return discrete subgraphs scored by prediction preservation: SubgraphX explores connected subgraphs by Monte-Carlo tree search, scoring each with a sampled Shapley reward
\citep{yuan2021subgraphx}, and Zorro greedily selects sparse masks with stability guarantees \citep{funke2023zorro}. GraphSVX \citep{duval2021graphsvx} and GraphSHAP-IQ
\citep{muschalik2024graphshapiq} bring Shapley values and exact any-order Shapley interactions to GNNs, with nodes and features as players, and \citet{covert2021explaining} unify such methods by their choice of removal operator, model behavior, and summary. \method instantiates that framework with cells as players and detachment on the complex as the removal operator. Shapley values over internal model components also precede us in neuron attribution \citep{ghorbani2020neuron} and structured pruning \citep{ancona2020shapleypruning}. \method-neighbor differs by scoring architectural neighborhoods with masked validation accuracy and by never selecting on the Shapley ranking itself (\cref{sec:framework-game2}). All of the above operate on nodes, edges, features, or network units. 

Beyond graphs, HyperEX explains hypergraph predictions through node--hyperedge associations \citep{maleki2023hyperex}, and SHypX returns explanatory subhypergraphs \citep{su2024shypx}. Because hypergraphs are simply rank-$\le 1$ combinatorial complexes, these approaches also do not handle cells across ranks. Counterfactual hypergraph explanations \citep{veglianti2026counterfactual} pursue prediction-changing edits, complementary to our prediction-preserving objective. FORGE \citep{sinha2025forge} lifts graphs to cell complexes before explaining, yet still outputs node-level attributions. \method instead defines games over cells across ranks of a combinatorial complex (of which graphs and hypergraphs are special cases) and over architectural neighborhoods. It is also the first to propose prediction attribution alongside validation-based architecture selection.
\section{Preliminaries}
\label{sec:preliminaries}

\method relies on three components: (i) \textit{the data}, combinatorial complexes (\cref{sec:prelim-cc}), whose
cells are the players of the cell game, (ii) \textit{the models}, topological neural networks (TNNs) specified by a set of neighborhoods (\cref{sec:prelim-gccn}), whose neighborhoods
are the players of the neighbor game, and (iii) \textit{the game-theoretic machinery}, Shapley values (\cref{sec:prelim-shapley}).

\subsection{Combinatorial Complexes and Their Cells}
\label{sec:prelim-cc}

Combinatorial complexes are discrete topological domains that generalize graphs, as follows.

\begin{definition}[Combinatorial complex and cells \citep{hajij2023tdl}]
\label{def:cc}
A \emph{combinatorial complex} is a triple $\mathcal{C}=(V, \mathcal{X}, \rk)$ where
$V$ is a finite set of nodes, $\mathcal{X} \subseteq
2^{V}\setminus\{\emptyset\}$ is a set of \emph{cells}, and $\rk :
\mathcal{X} \to \mathbb{Z}_{\ge 0}$ is a rank function monotone with
respect to inclusion: $x \subseteq y \implies \rk(x) \le \rk(y)$.
\end{definition}

\textit{Cells} are the building blocks of the combinatorial complex (depicted in \cref{fig:overview}, top left). Cells of rank $0$, $1$ and $2$ are called nodes (depicted in blue), edges (depicted in pink) and faces (depicted in burgundy), respectively. Cells of higher ranks represent higher multi-way interactions between nodes. Graphs and hypergraphs are special cases of combinatorial complexes, where all cells have rank $\leq 1$. Other higher order domains used in TDL are also special cases, such as simplicial complexes, whose rank-2 cells are triangles and whose cells are closed under taking subsets, and cellular complexes, whose faces may be arbitrary polygons. Combinatorial complexes generalize them all \citep{hajij2023tdl}, which is why we define \method on them.

Raw data is often provided in graph format. A topological \emph{lifting} converts a graph into a higher order domain by promoting substructures of nodes to higher-order cells. For example a cycle of nodes in a graph can be defined as a face in a cellular complex \citep{bodnar2021cwn}. Of the datasets we use, only MANTRA
\citep{ballester2025mantra} is natively defined on a higher order domain, namely the simplicial complex. We lift the other (graph) datasets we use to the cellular complex domain using liftings available in TopoBench \citep{telyatnikov2024topobench}. We will see that the choice of lifting
impacts both accuracy and explanation quality
(\cref{sec:predictions}).

\subsection{Topological Neural Networks and Their Neighborhoods}
\label{sec:prelim-gccn}

TNNs generalize GNNs to consider higher order domains. Here, we specifically consider the combinatorial complex for its generality, encompassing graph, hypergraphs, simplicial/cell complexes, and more. A TNN architecture is specified by a set of \emph{neighborhoods}, the relationships through which cells interact on the combinatorial complex. We first define neighborhoods formally, then provide preliminaries for the two model families we use in experiments: GCCNs, which pass messages along the chosen neighborhoods, and HOPSE, which encodes them without message passing.

\begin{definition}[Neighborhood \citep{hajij2023tdl,hopse2025}]
\label{def:neighborhood}
A neighborhood on a combinatorial complex $\mathcal{C}=(V,\mathcal{X},\rk)$ is a function $N : \mathcal{X} \to 2^{\mathcal{X}}$ assigning to each cell $x$ a set $N(x)$ of cells, its neighbors. The standard neighborhoods are indexed by ranks. For a cell $x$ of rank $t$ and a rank $s \neq t$, the \emph{incidence}
\begin{equation*}
I_{s \to t}(x) \;=\; \{\, y \in \mathcal{X} \mid \rk(y) = s,\; x \subsetneq y \text{ or } y \subsetneq x \,\}
\end{equation*}
collects the rank-$s$ cells that contain $x$ (when $s > t$) or are contained in $x$ (when $s < t$), and the \emph{adjacency}
\begin{equation*}
A_{t,s}(x) \;=\; \{\, y \in \mathcal{X}\setminus\{x\} \mid \rk(y) = t,\; I_{s \to t}(x) \cap I_{s \to t}(y) \neq \emptyset \,\}
\end{equation*}
collects the rank-$t$ cells that share a rank-$s$ incident cell with $x$. 
\end{definition}

Intuitively, neighborhoods are the routes along which cells exchange information. Incidences connect cells of different ranks, \eg $I_{1 \to 0}$ links each node to the edges containing it; adjacencies connect cells of the same rank through a shared cell, \eg $A_{0,1}$ links nodes joined by an edge. In practice each neighborhood is stored as a sparse matrix over cells. 

 Given a set $\mathcal{N} = \{N_1, \dots, N_m\}$ of neighborhoods, each $N \in \mathcal{N}$ decomposes $\mathcal{C}$ into an ordinary graph on cells, the \emph{strictly augmented Hasse graph} $\mathcal{G}_N$, where cells participating in the neighborhood are nodes and the relationships of the neighborhood are edges \citep{papillon2025topotune}. With this formulation, the combinatorial complex $\mathcal{C}$ can be represented as a collection of smaller graphs, one per neighborhood. Both TNN families we use are built directly on this collection. 
 Because both models represent each neighborhood as an architectural component, removing a neighborhood from $\mathcal{N}$ is a well defined operation on the model. 
 
\begin{definition}[Generalized combinatorial complex network (GCCN) \citep{papillon2025topotune}]
\label{def:gccn}
A GCCN stacks $L$ message-passing layers over the strictly augmented Hasse graphs $\{\mathcal{G}_N\}_{N \in \mathcal{N}}$. At layer $\ell$, every cell $x$ exchanges a message with its neighbors along each graph $\mathcal{G}_N$, using a GNN layer $\psi_N$ dedicated to that neighborhood; the messages arriving from a single neighborhood are pooled together, and the results are then pooled once more across neighborhoods to update $x$'s feature:
\begin{equation}
\label{eq:gccn-layer}
\mathbf{h}^{(\ell+1)}_x
  \;=\;
  \phi\!\Bigl(\mathbf{h}^{(\ell)}_x,\;
  \bigotimes_{N \in \mathcal{N}}\;
  \bigoplus_{y \in N(x)}
  \psi_{N}\bigl(\mathbf{h}^{(\ell)}_x, \mathbf{h}^{(\ell)}_y\bigr)\Bigr).
\end{equation}
\end{definition}
 
\begin{definition}[Higher-Order Positional and Structural Encoder (HOPSE) \citep{hopse2025}]
\label{def:hopse}
HOPSE replaces every GNN layer $\psi_N$ with a fixed positional or structural encoder, such as Laplacian eigenvectors, computed once on the strictly augmented Hasse graph $\mathcal{G}_N$ rather than learned during training. For every cell $x$, HOPSE collects the encodings $x$ receives this way from each of its Hasse graphs, concatenates them with $x$'s own input feature, and passes the result through a small learnable network, shared across cells of the same rank, to produce $x$'s final feature. %
\end{definition}

\subsection{Measuring a Contribution to a Game: Shapley Values and Interactions}
\label{sec:prelim-shapley}

A \emph{cooperative game} $(P, v)$ consists of players $P$, $|P| = n$, and a
value function $v : 2^{P} \to \mathbb{R}$ with $v(\emptyset) = 0$, where $v(S)$
is the gain that the \emph{coalition} $S \subseteq P$ secures on its own. The
question is how to divide the total gain $v(P)$ among the $n$ players.

The \emph{Shapley value}
\citep{shapley1953value} of player $i$ averages its marginal
contribution over all coalitions:
\begin{equation}
\label{eq:shapley}
\phi_i(v) \;=\; \sum_{S \subseteq P \setminus \{i\}}
  \frac{|S|!\,(n - |S| - 1)!}{n!}
  \bigl[v(S \cup \{i\}) - v(S)\bigr].
\end{equation}
It is the unique attribution satisfying four desirable axioms: (i) \emph{efficiency}, i.e., attributions credit
the total gain exactly, $\sum_i \phi_i = v(P)$, (ii) \emph{symmetry} where $\phi_i = \phi_j$
whenever $v(S \cup \{i\}) = v(S \cup \{j\})$ for all $S$, (iii) \emph{null player}
where $\phi_i = 0$
whenever $v(S \cup \{i\}) = v(S)$ for all $S$, and (iv) \emph{linearity} which makes credit
additive across games, $\phi_i(v + w) = \phi_i(v) + \phi_i(w)$. These axioms
explain the adoption of Shapley values in explainable machine learning. For this reason, this is the attribution that we will use in \method.

\section{Method: TopoSHAP}
\label{sec:framework}

\method asks two questions about a trained TNN. Which cells of an input drove its prediction? Which neighborhoods of the architecture drove its performance? Each question can be answered with a cooperative game (\cref{sec:prelim-shapley}), which we call \method-cell and \method-neighbor, respectively. \cref{fig:overview} compares the two games.

\subsection{\method-cell: Explaining Predictions.}
\label{sec:framework-game1}

\paragraph{The cell game.} The players $P$ are the cells of the input complex, at every rank; $n = |P|$. A coalition $S \subseteq P$ specifies a perturbed input $\mathcal{C}[S]$ in which every cell outside $S$ is \emph{detached}. Specifically, its feature row is zeroed, neighborhoods are rebuilt without it, and it is excluded from readout pooling. Retained cells keep the input features they were supplied with; cells are independent players, so a face remains when one of its vertices is detached, losing only its incidence to that node. For HOPSE, whose inputs include precomputed encodings of the complex, the encodings are recomputed on $\mathcal{C}[S]$. 

The value of a coalition is $v(S) = f(\mathcal{C}[S]) - f(\mathcal{C}[\emptyset])$, where $f(\mathcal{C}[S]) $ is the model's prediction given the coalition, and $f  (\mathcal{C}[\emptyset])$ is the model's prediction given an empty complex, leaving the encoder and readout biases. At $S = P$ the rebuilt input reproduces the model's stored forward pass. \Cref{app:cell-intervention} gives the full algorithm.

\paragraph{Output 1: cell attributions.} The Shapley value $\phi_i(v)$ of \cref{eq:shapley} credits each player with its average marginal contribution to the coalition value $v(S)$. The Shapley axioms ensure efficiency, meaning the Shapley values sum to $v(P)$, as well as null-player behavior, meaning a player \textit{j} the model never uses gets $\phi_j(v)=0$.
To compute exact Shapley values, \cref{eq:shapley} requires every possible coalition. While this is feasible for small numbers of players (such as in \method-neighbor where $n = \ResNumNeighborhoods$ so $2^{n} = 2048$ coalitions), it is prohibitively expensive for many players (such as in \method-cell for large complexes). In that case, we follow \citet{castro2009polynomial} and draw $T$ random orderings of the players, add them one at a time to a coalition, and average the marginal changes. This unbiased estimate costs $O(Tn)$ model evaluations. 

\paragraph{Output 2: explanation via greedy pruning.} An explanation is a small set of players that together preserve the model's prediction; we approximate it with one greedy candidate per size. Starting from all players $P$, we repeatedly eliminate the player whose removal lowers $v$ the least. Because the steps are nested, one pass returns a candidate coalition $\hat{S}_{b}$ for every size $b$, at most $(n^{2}+n)/2$ evaluations in total. Each $\hat{S}_{b}$ is a compact, high-scoring coalition on the greedy path. Greedy elimination carries no optimality guarantee, so $\hat{S}_{b}$ need not be the best size-$b$ coalition. 

\paragraph{Axioms as unit tests.} For an exact game the Shapley axioms hold with equality; our implementation matches brute force to \ResExactnessMaxErr, and separate consistency checks test the masking itself (\cref{app:cell-intervention}).

\subsection{\method-neighbor: Explaining Performance}
\label{sec:framework-game2}
\label{sec:method-pipeline}

\method-neighbor evaluates how neighborhoods contribute to a trained model's validation accuracy. These evaluations support both neighborhood attribution and architecture selection. The players are the candidate neighborhoods $\mathcal{N}$, with $n=|\mathcal{N}|$. Our main rank-2 architecture-selection experiments consider eleven neighborhoods. Here, a coalition $S \subseteq \mathcal{N}$ specifies a model where the neighborhood-specific modules (message-passing modules in GCCN or encoding MLP blocks in HOPSE) are constrained to neighborhoods in $S$. The encoder and readout backbones are left untouched.

\paragraph{Evaluating coalitions.}
In practice, training a separate model for every coalition requires $2^n$ training runs. With the goal of \method-neighbor being practical in applied TDL settings, we propose a cheaper alternative. Specifically, we briefly train a full version of the model containing all neighborhoods, and then evaluate coalitions on a frozen version of it. For each $S$, we mask any neighborhood-specific modules outside $S$ and measure the resulting validation accuracy, $a(S)$. We define the value function $v(S)=a(S)-a(\emptyset)$ where $a(\emptyset)$ is the accuracy of the same model with all neighborhood components masked, leaving its encoder and readout active. Each coalition therefore requires one validation pass. This brief initial training follows prior work on identifying ideal architectures from early epochs \citep{you2020earlybird,frankle2020earlyphase} and low-fidelity candidate ranking in architecture and hyperparameter search \citep{li2018hyperband,mellor2021naswot}.

\paragraph{Output 1: neighborhood attributions.}
The Shapley value of a neighborhood measures its average marginal contribution to $v(\mathcal{N})$. We compute these values exactly by evaluating all $2^n$ coalitions $S \subseteq \mathcal{N}$. We note that these attributions describe neighborhood contributions in one frozen model, meaning they do not establish which neighborhoods a separately trained model would need. For example, a model may learn to rely on one of two neighborhoods carrying equally useful information. 

\paragraph{Output 2: architecture selection.}
We use the coalition evaluations to select architectures through backward elimination, following backward stepwise selection and recursive feature elimination \citep{guyon2002rfe,hastie2009esl}. Starting from $S=\mathcal{N}$, we evaluate every possible single-neighborhood removal and retain the coalition with the highest validation accuracy. We repeat this process down to one neighborhood, obtaining one candidate per size. Each step reassesses removal effects given the neighborhoods that remain (\cref{alg:neighbor-selection}). 

From these $n$ candidates, we choose $B$ to train further, where $1 \leq B \leq n$. We first select the \emph{stop candidate}: the architecture immediately before the first removal whose accuracy loss exceeds
\[
\tau=\sqrt{\frac{p(1-p)}{m}},
\qquad p=a(\mathcal{N}),
\]
where $m$ is the number of validation examples and $p$ is the briefly trained full model's validation accuracy. This estimated binomial standard error provides a data-dependent pruning threshold. If no removal exceeds it, the stop candidate is the last remaining neighborhood. The threshold marks a candidate without halting elimination, and does not bound the cumulative accuracy loss relative to the full model.

We select the remaining $B-1$ architectures by selecting the remaining candidates with the highest validation accuracy. Each of the $B$ selected architectures is initialized using the full frozen model and then trained to completion. Of the $B$ trained models, we select the one with the highest final validation accuracy. This pipeline provides an inexpensive alternative to proposing architectures based on early neighborhood contributions. \Cref{alg:neighbor-selection} gives the complete procedure.

\section{Results I: Explaining Predictions with \method-cell}
\label{sec:predictions}

This section asks which cells a trained TNN relies on, considering benchmark explanation graph datasets, lifted topological versions of those datasets, as well as unlabeled standard topological datasets.

\textbf{Setup.} We use four GraphXAI molecular benchmark data with ground truth explanations from \citet{agarwal2023graphxai}, four lifted topological benchmark data from TopoBench \citep{telyatnikov2024topobench}, and the MANTRA dataset \citep{ballester2025mantra}, a native simplicial complex dataset. All comparisons use identical splits and three seeds (mean $\pm$ sd).  Explanation accuracy is Graph Explanation Accuracy (GEA), the Jaccard overlap between a size-$b$ explanation and a same-size ground-truth motif \citep{agarwal2023graphxai}. In cases where no ground truth is available, we measure explanation faithfulness using deletion-Area-Under-the-Curve \citep[deletion-AUC;][]{petsiuk2018rise}. To compute it, cells are detached cumulatively from most to least important and the predicted-class probability is integrated over the removal process. A faithful ranking collapses the prediction quickly, so lower is better (formula in \cref{app:cell-intervention}). Lastly, the \emph{alignment gap} $\Delta = v(\hat{S}_{b}) - v(S_{\mathrm{gt}})$ compares the greedy candidate $\hat{S}_{b}$ to the annotated ground truth motif $S_{\mathrm{gt}}$. A positive $\Delta$ occurs when a same-size coalition scored above the annotated motif, indicating that the annotation is in fact not the game's optimum on that instance. Datasets and training details are in \cref{app:setup}; evaluation settings in \cref{app:cell-results}.

\subsection{\method-cell on Graphs Matches GNN Explainers for Cheap}
\label{sec:predictions-rank1}

\begin{table}[t]
\centering
\small
\setlength{\tabcolsep}{3pt}
\caption{\textbf{Explanation accuracy on graphs.} GEA on parameter-matched GNN models; per method the better of GCN/GIN is shown (both in \cref{app:tables}). Last column: measured seconds per instance; $^{\dagger}$plus one-time training (142--1501\,s per dataset). Bold is best and gray is within $\leq 1 \sigma$.}
\label{tab:graph-benchmarks}
\begin{tabular}{lccccc}
\toprule
 & Benzene & FluorideCarbonyl & Mutagenicity & AlkaneCarbonyl & s / inst. \\
\midrule
\method{} (attribution) & 0.859 $\pm$ 0.009 & \cellcolor{black!10}0.507 $\pm$ 0.047 & \cellcolor{black!10}\textbf{0.798 $\pm$ 0.124} & \cellcolor{black!10}\textbf{0.500 $\pm$ 0.081} & 3.3--4.0 \\
\method{} (explanation) & \cellcolor{black!10}\textbf{0.969 $\pm$ 0.009} & 0.340 $\pm$ 0.010 & 0.480 $\pm$ 0.135 & 0.354 $\pm$ 0.125 & 0.33--0.67 \\
SubgraphX & 0.796 $\pm$ 0.029 & 0.439 $\pm$ 0.107 & 0.344 $\pm$ 0.036 & 0.296 $\pm$ 0.010 & 21--52 \\
PGExplainer & 0.642 $\pm$ 0.044 & \cellcolor{black!10}\textbf{0.555 $\pm$ 0.067} & 0.290 $\pm$ 0.102 & 0.205 $\pm$ 0.160 & $<$0.01$^{\dagger}$ \\
GNNExplainer & 0.443 $\pm$ 0.054 & 0.210 $\pm$ 0.014 & 0.158 $\pm$ 0.023 & 0.162 $\pm$ 0.049 & 0.80--0.82 \\
Random & 0.250 $\pm$ 0.000 & 0.219 $\pm$ 0.000 & 0.132 $\pm$ 0.000 & 0.173 $\pm$ 0.000 & $<$0.01 \\
\bottomrule
\end{tabular}
\end{table}

\expclaim{\method-cell matches within $1\sigma$ or outperforms every baseline.} In \cref{tab:graph-benchmarks}, all methods explain the same parameter-matched models of two types: GCN \citep{kipf2017semisupervised} and GIN \citep{xu2018how}. The table shows each method's better model-type. Per-model-type results are in \cref{tab:by-subject}. We evaluate both outputs by their overlap with the annotated motif (the ground truth): attribution selects the $b$ highest-scoring nodes, while greedy pruning retains a size-$b$ set through successive removals. Greedy pruning achieves higher motif overlap on Benzene with GIN; attribution-based selection performs better on Mutagenicity and Alkane-Carbonyl for both model types.


\expclaim{\method's explanation output is 64--112$\times$ cheaper per instance than search-based explanation.} The explanation output takes 0.33--0.67\,s per instance and the attribution output 3.3--4.0\,s, against 21--52\,s for its graph counterpart SubgraphX  (\cref{tab:graph-benchmarks}). PGExplainer has a costly training setup: for computing explanations it exceeds \method's cost on every dataset (\cref{tab:runtime}). 

\subsection{\method-cell on Higher-Order Data Recovers Explanations}
\label{sec:predictions-lifted}
\label{sec:predictions-higher}

We consider the rank-2 GCCN containing all eleven possible neighborhoods. We lift the GraphXAI datasets conventionally by converting chordless cycles of up to 10 nodes to a face (\cref{app:datasets}).

\expclaim{Standard lifting preserves aggregated node information.}
The standard lifting assigns face features using the sum of their incident nodes' features. In \method-cell, because retained cells keep their original features during masking, this summary information remains available regardless of node or edge masking. We find that selected coalitions sometimes retain faces while discarding edges (Benzene, \cref{app:lifting-diagnostics}), suggesting some models use these summaries. The selected coalitions also outscore the annotated motifs on most evaluated instances ($\Delta > 0$; \cref{tab:lifted}), showing that motif recovery and the model's preferences under masking can diverge.

\expclaim{Constant higher-rank features improve motif recovery
on three of four datasets.}
Building on our finding of summary information caching structural information, we devise a new lifting where every higher-rank feature is instead the same constant all-ones vector. Under this ``structure-only'' lifting, greedy explanations achieve higher mean GEA on three datasets and attribution results are mixed (\cref{tab:lifted}). Mean predictive accuracy also increases on three datasets. We consider models trained on Benzene as a specific example. Moving from standard to structure-only lifting increases face retention from $0.84 \pm 0.14$ to $0.95 \pm 0.05$ and boundary-edge recovery from $0.40 \pm 0.14$ to $0.86 \pm 0.06$
(\cref{app:lifting-diagnostics}). Thus, explanations more often retain both the face and its boundary. More broadly, these findings support the importance of feature lifting
for explanation recovery. %

\begin{table}[t]
\centering
\small
\setlength{\tabcolsep}{3pt}
\caption{\textbf{Explanation accuracy on lifted complexes.} Node-projected GEA for both \method{} outputs of the GCCN with all eleven neighborhoods under the two liftings; the lower block gives each model's test accuracy and its share of candidates with $\Delta > 0$. Bold is best and gray is within $\leq 1 \sigma$.}
\label{tab:lifted}
\begin{tabular}{lcccc}
\toprule
 & Benzene & FluorideCarbonyl & Mutagenicity & AlkaneCarbonyl \\
\midrule
attribution, structural lifting & 0.471 $\pm$ 0.018 & \cellcolor{black!10}0.491 $\pm$ 0.035 & 0.326 $\pm$ 0.065 & \cellcolor{black!10}0.333 $\pm$ 0.097 \\
attribution, standard lifting & 0.459 $\pm$ 0.192 & \cellcolor{black!10}\textbf{0.516 $\pm$ 0.089} & 0.257 $\pm$ 0.077 & \cellcolor{black!10}0.348 $\pm$ 0.043 \\
explanation, structural lifting & \cellcolor{black!10}\textbf{0.725 $\pm$ 0.105} & 0.426 $\pm$ 0.027 & \cellcolor{black!10}\textbf{0.436 $\pm$ 0.081} & \cellcolor{black!10}0.362 $\pm$ 0.130 \\
explanation, standard lifting & 0.509 $\pm$ 0.188 & 0.372 $\pm$ 0.004 & 0.289 $\pm$ 0.117 & \cellcolor{black!10}\textbf{0.364 $\pm$ 0.053} \\
\midrule
model accuracy, structural lifting & 0.942 $\pm$ 0.005 & 0.947 $\pm$ 0.003 & 0.976 $\pm$ 0.002 & 0.967 $\pm$ 0.006 \\
model accuracy, standard lifting & 0.928 $\pm$ 0.003 & 0.936 $\pm$ 0.003 & 0.969 $\pm$ 0.009 & 0.971 $\pm$ 0.005 \\
share with $\Delta > 0$, structural lifting & 0.74 & 0.93 & 0.97 & 0.99 \\
share with $\Delta > 0$, standard lifting & 0.97 & 0.94 & 1.00 & 0.99 \\
\bottomrule
\end{tabular}
\end{table}

\subsection{\method-cell Is the Most Faithful Explainer Without Annotations}
\label{sec:predictions-nogt}

\expclaim{\method-cell beats occlusion, gradient-times-input, and random on every unannotated dataset.} Here we apply \method-cell to benchmark topological datasets hosted in TopoBench \citep{telyatnikov2024topobench}. Because there is no ground truth annotation, we score explanations against the model's own prediction by deletion-AUC (\cref{tab:nogt}; protocol in \cref{app:faithfulness}). We hypothesize that \method outperforms other baselines by attributing cells across coalitions rather than in isolation.

\expclaim{Rank-specific attributions help explain models'
prediction strategies.}
Here, we consider MANTRA \citep{ballester2025mantra}, the natively topological dataset, and its orientability classification task. HOPSE performs best overall (AUROC 0.93), followed by GCCN (0.79) and the published TopoBench baseline (0.52). However, when we restrict evaluation to surfaces with matched cell counts, GCCN falls to near chance (0.51), while HOPSE retains 0.79. Thus, HOPSE can distinguish orientability using information beyond the numbers of nodes, edges, and triangles (protocol in \cref{app:mantra}).  Cell attributions help interpret this difference. Consider an orientable torus and a non-orientable Klein bottle with identical cell counts (\cref{fig:app-mantra}). Attributions show GCCN leverages triangles, but fails to distinguish triangles of orientable and the non-orientable surfaces. HOPSE instead concentrates its credit on nodes and edges, but successfully assigns attributions of opposite sign. We hypothesize that HOPSE succeeds because its spectral encodings provide global information about the complex's topology. Giving GCCN access to these encodings supports the hypothesis: with Hodge Laplacian eigenvectors concatenated to its input features, GCCN reaches a balanced accuracy of 0.88, exceeding HOPSE's 0.79 (\cref{tab:mantra-features}). This is an example of how \method-cell can shed light on models' inner workings and help design better TNNs for specific tasks.

\section{Results II: Explaining Performance with \method-neighbor}
\label{sec:performance}

\subsection{Architecture Selection in Practice}
\label{sec:performance-selection}

\begin{figure}[!t]
  \centering
  \includegraphics[width=\linewidth]{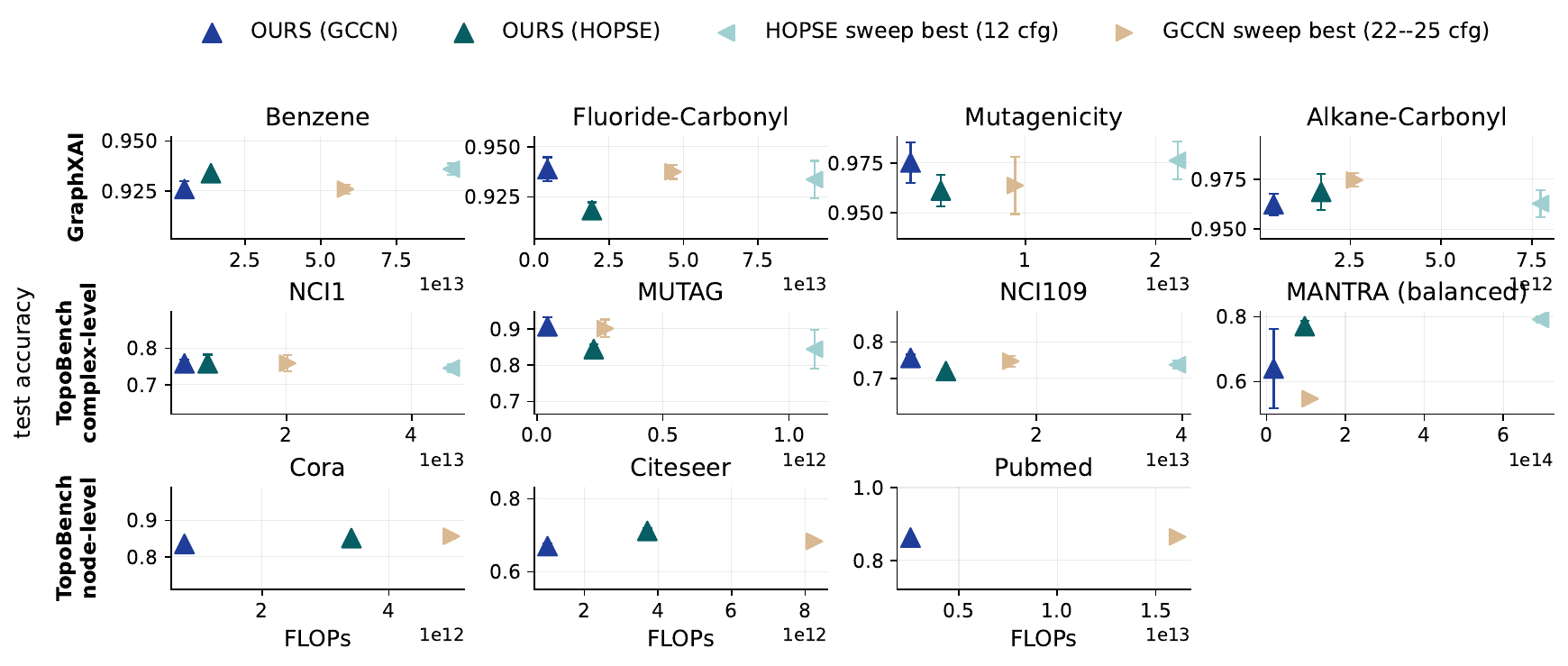}
  \vspace{-8mm}
  \caption{\textbf{Architecture selection: accuracy versus total compute.} Dark markers denote \method-neighbor; pale markers denote each family's validation-selected sweep winner, charged for the full sweep. Error bars show standard deviation over five seeds for GCCN $B=2$ and three otherwise. Due to its practical costs, HOPSE sweeps are limited to complex-level datasets and we avoid training any HOPSE models on Pubmed. MANTRA uses balanced test accuracy.}
  \label{fig:lead}
\end{figure}

\method-neighbor reduces total training compute by a factor of 4.5--11.3 relative to the evaluated architecture sweeps. The resulting accuracy--compute trade-off varies by dataset, as shown in \cref{fig:lead,tab:performance-summary}. We evaluate GCCN and HOPSE across 11 datasets against their respective reported architecture search grids \citep{papillon2025topotune,hopse2025}. Our pipeline uses a five-epoch full model to select $B=2$ smaller architectures, completes the training of these smaller architectures, and selects one by validation accuracy. Reported costs include search, candidate training, and preprocessing (\cref{app:flops-conventions}). Controls using the same initialization and continuation settings compare elimination with a random two-candidate shortlist and the full-model default (\cref{tab:shortlist-baselines}). We find that elimination generally improves mean accuracy.

\subsection{Controlled Evaluation with Exhaustive Training}
\label{sec:performance-controlled}

We separately train all $2^6=64$ subsets of six neighborhoods across twelve GCCN dataset/hyperparameter settings and three seeds (\cref{app:benchmark}). This exhaustive retraining allows us to compute the neighborhoods' Shapley values from separately trained coalitions and compare them with attributions from the masked five-epoch model; the selection pipeline of \cref{sec:performance-selection} itself never consumes attributions, scoring candidate removals directly by masked coalition accuracy.

\paragraph{Masked attributions describe reliance in the trained model.}
Masked-model and retrained-model Shapley values show weak pooled agreement ($\rho=0.23$), rising to $0.32$ when masking is scored by accuracy rather than logits (\cref{tab:attr-agreement}). Masking uses validation data, whereas retraining is scored by test accuracy or negative MAE, so this comparison also reflects differences in data and scoring (\cref{app:attr-agreement}). The results caution against interpreting fixed-model attributions as necessity after retraining. 

\paragraph{Backward elimination generates competitive candidates in an oracle analysis.}
We score removals using the separately trained coalitions' seed-averaged test performance. On five held-out settings across four datasets, the cheap greedy architecture selection (outlined in \cref{sec:framework-game2}) contains the exhaustive benchmark's best architecture in four settings (\cref{tab:oracle-generation}). This is encouraging evidence that our pipeline can construct competitive candidate sequences (\cref{app:selection-evidence}).

\section{Limitations and Conclusion}
\label{sec:limitations}


\method brings prediction and architecture explanations into a unified explainability framework for Topological Deep Learning, providing insight into model predictions and performance. We conclude by describing two limitations which are also promising avenues for future work. First, scaling to larger complexes requires more efficient coalition evaluation. Although greedy pruning is substantially cheaper than SubgraphX on our benchmarks
(\cref{tab:graph-benchmarks}), it requires $O(n^{2})$ model evaluations for $n$ cells, while attribution requires $O(Tn)$ evaluations for $T$ sampling passes. Truncated or sampled search offers a path toward complexes with thousands of cells. Second, our evaluation against ground truth explanations remains graph-based in this work. Developing explanation-annotated benchmarks on general combinatorial complexes, building on graph benchmarks and emerging synthetic hypergraph benchmarks \citep{su2024shypx}, will be paramount for developing better explanations for, and thus more trustworthy, Topological Deep Learning.


\subsection*{Acknowledgments}
M. P. and N. M. acknowledge partial support from NSF grant 2602079 and the Noyce foundation. M. P. and N. M. also thank Goodfire AI for providing access to Silico through its academic research grant program.
M. P. also acknowledges support from NSF Career grant 2240158 and the National Science and Engineering Research Council of Canada. 

\subsection*{AI use statement}

In this work, we used generative AI tools to assist with implementing and debugging the proposed methods, running experiments, and drafting initial versions of portions of the manuscript. We also used generative AI tools to assist with summarizing existing literature; all cited literature and associated claims were independently reviewed by the authors. AI tools were not used to conceptualize the project, develop its core research ideas or methodology or interpret the experimental results. All datasets used in this work were publicly available prior to this study and were not generated using AI. AI-assisted code was reviewed, modified, and tested by the authors, and all experimental results and analyses were independently verified by the authors. The authors wrote a substantial portion of the main text and heavily revised and edited all AI-assisted text. Appendix text is a more equal combination of author and AI written text. We take full responsibility for the final content of the paper, including all claims, analyses, code, and artifacts.

\subsection*{Reproducibility statement}
Datasets, splits, and all model, training, and explainer settings are specified in \cref{app:setup}; evaluation protocols and full results in \cref{app:cell-results,app:neighbor-results}; the software environment in \cref{app:repro}. The implementation is made available at \href{https://github.com/mathildepapillon/toposhap}{github.com/mathildepapillon/toposhap}.

\bibliographystyle{plainnat}
\bibliography{references}

\clearpage
\appendix
\section{Additional Method Details}
\label{app:method}

\subsection{Cell Masking and Explanation Algorithms}
\label{app:cell-intervention}

\textbf{Masked input.} Given a complex $\mathcal{C}$ with cells $P$ and a
coalition $S \subseteq P$, the masked input $\mathcal{C}[S]$ is built by
detaching every cell outside $S$: its feature row is zeroed, its entries
are removed from every neighborhood matrix in which it appears (as
source, target, or intermediate cell, with composed neighborhoods
rebuilt from the incidences of the remaining cells, using the same
constructors that built the original input), and it is excluded from
readout pooling. Retained cells keep the input features they were
supplied with; because cells are independent players, a face can remain
after one of its nodes is detached, so $\mathcal{C}[S]$ is a masked
complex and not in general a node-induced subcomplex. For HOPSE, whose
inputs include precomputed positional and structural encodings, the
encodings are recomputed on each $\mathcal{C}[S]$. On plain graphs we
take the nodes as players and let incident edges follow their
endpoints, in which case $\mathcal{C}[S]$ is the node-induced subgraph.

\textbf{Target score.} $f$ reads one fixed class logit, held fixed
across all coalitions of an instance: the annotated class on labeled
benchmarks, the predicted class on datasets without annotations (no
labels consulted), and the orientable-class logit margin on MANTRA.
Coalition values are $v(S) = f(\mathcal{C}[S]) -
f(\mathcal{C}[\emptyset])$, where $\mathcal{C}[\emptyset]$ detaches
every cell, leaving the encoder and readout biases.

\begin{figure}[t]
  \centering
  \includegraphics[width=\linewidth]{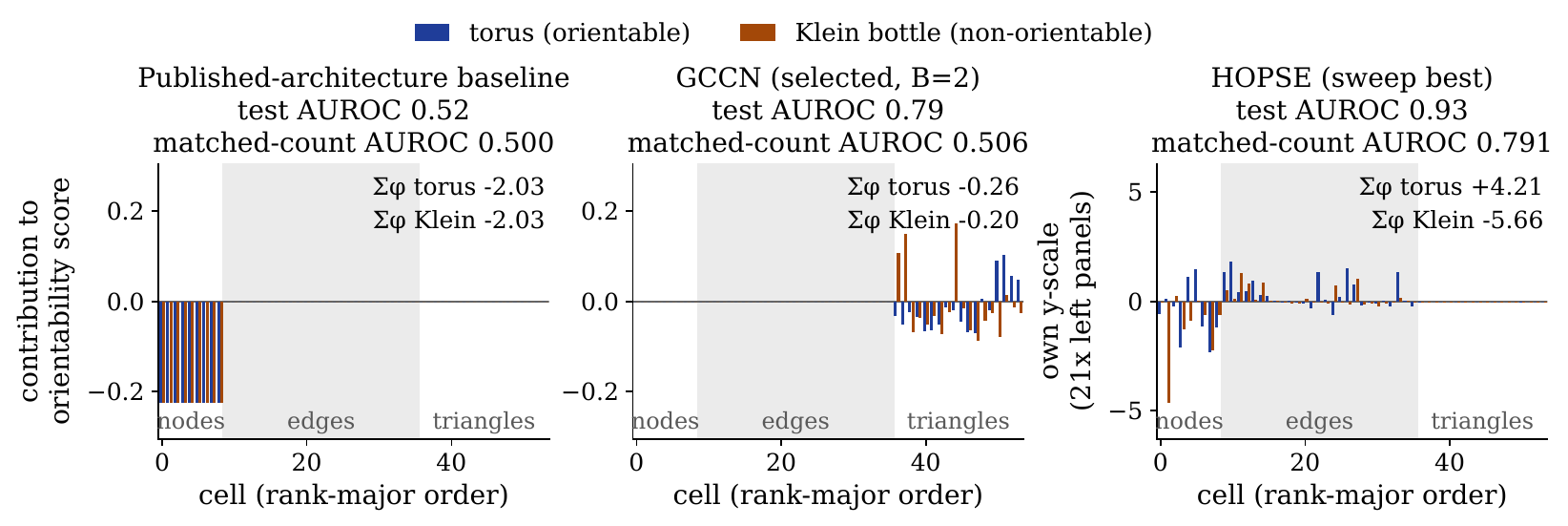}
  \caption{\textbf{MANTRA attributions on a count-matched pair.} One
    orientable torus and one non-orientable Klein-bottle-family
    surface with identical cell counts (9 nodes, 27 edges, 18
    triangles), drawn from the smallest count group containing both
    classes: per-cell contribution to the orientable-class logit
    margin for the three models of \cref{sec:predictions-nogt}
    (published-architecture baseline; the GCCN selected by
    \method-neighbor at $B{=}2$; the sweep-best HOPSE). Panel headers
    give each model's overall and count-conditional test AUROC. HOPSE
    is drawn on its own $y$-scale, stated on the axis. Per-cell values
    are single sampled-game estimates at $T = 512$ orderings
    (\cref{app:cell-intervention}); no aggregation, hence no error
    bars.}
  \label{fig:app-mantra}
\end{figure}

\textbf{Attribution (Output 1).} Exact Shapley values enumerate all
$2^{n}$ coalitions when $n$ is small. Otherwise we sample player
orderings, add players in the sampled order, and average the marginal
changes \citep{castro2009polynomial}. The budget $T$ counts every
evaluated ordering: orderings are drawn uniformly at random and each
drawn ordering is followed by its reverse, so $T$ orderings comprise
$T/2$ random draws and their reversals. A reversed ordering visits its
own coalitions; all coalition values are cached by bitmask, so the
total cost is at most $Tn + 1$ model evaluations and fewer in
practice. Explanation experiments use $T = 128$ with estimator seed 0,
distinct from the training seeds; the per-cell estimates of
\cref{fig:app-mantra} use $T = 512$.

\textbf{Explanation (Output 2).} Greedy backward pruning
(\cref{alg:cell-game}) starts from $P$ and repeatedly removes the cell
whose removal lowers $v$ the least, producing one nested candidate
$\hat{S}_{b}$ per size $b$ in at most $(n^{2}+n)/2$ evaluations. The
candidates are compact, high-scoring coalitions on the greedy path and
carry no optimality guarantee.

\begin{algorithm}[H]
\caption{\method-cell on one instance}
\label{alg:cell-game}
\begin{algorithmic}[1]
\Require complex $\mathcal{C}$ with cells $P$, $n = |P|$; model logit
         $f$; sampling budget $T$
\State define $v(S) = f(\mathcal{C}[S]) - f(\mathcal{C}[\emptyset])$,
       cached by coalition
\State \textbf{Output 1:} $\phi \gets$ exact Shapley values if $2^{n}$
       is affordable, else the $T$-ordering estimate (each drawn
       ordering followed by its reverse)
\State \textbf{Output 2:} $S \gets P$;\quad $\hat{S}_{n} \gets P$
\For{$b = n-1, \ldots, 1$}
    \State $i^{*} \gets \arg\max_{i \in S} v(S \setminus \{i\})$,
           ties toward the lower cell index
    \State $S \gets S \setminus \{i^{*}\}$;\quad $\hat{S}_{b} \gets S$
\EndFor
\State \Return $\phi$ and $\{\hat{S}_{b}\}_{b=1}^{n}$
\end{algorithmic}
\end{algorithm}

\textbf{Implementation checks.} At $S = P$ the rebuilt input must
reproduce the model's stored forward pass exactly; this check runs at
construction, and the HOPSE variant additionally checks that recomputed
full-coalition encodings reproduce the stored prediction within
tolerance. The Shapley computation is verified against brute-force
enumeration to \ResExactnessMaxErr on small games. These checks verify
the implementation of the specified game; separate consistency checks
(attribution invariance under batch permutation, equality on provably
symmetric cells) test the masking itself.

\subsection{Neighborhood Masking and Architecture Selection}
\label{app:selection-pipeline}
\label{app:pipeline-protocol}

\textbf{The eleven neighborhoods.} Every subset mask indexes
neighborhoods in the order listed below, with the code label followed
by the notation of \cref{def:neighborhood} ($I_{s \to t}$: messages
from rank-$s$ cells to their incident rank-$t$ cells):
up\_adjacency-0 $= A_{0,1}$, up\_incidence-0 $= I_{0 \to 1}$,
down\_incidence-1 $= I_{1 \to 0}$, up\_adjacency-1 $= A_{1,2}$,
down\_incidence-2 $= I_{2 \to 1}$, 2-up\_adjacency-0 $= A_{0,2}$,
down\_adjacency-1 $= A_{1,0}$, up\_incidence-1 $= I_{1 \to 2}$,
down\_adjacency-2 $= A_{2,1}$, 2-up\_incidence-0 $= I_{0 \to 2}$,
2-down\_incidence-2 $= I_{2 \to 0}$. The exhaustive benchmark of
\cref{app:benchmark} uses the first six.

\textbf{Masking a neighborhood.} In a GCCN, each neighborhood owns one
message-passing module per layer; in HOPSE, one encoding block (the
fixed encoder on that neighborhood's graph plus its learned MLP).
Masking a neighborhood silences its modules; removing it deletes them,
and for HOPSE also removes that block's preprocessing, which is charged
accordingly (\cref{app:flops-conventions}).

\textbf{Selection.} \Cref{alg:neighbor-selection} restates the
procedure of \cref{sec:framework-game2}. The five-epoch initial model
defines $a(S)$, the masked validation accuracy; backward elimination
removes, at each step, the neighborhood whose removal keeps $a$
highest (ties toward the first neighborhood in the order listed
above), continuing down to one neighborhood and yielding one candidate
per size. The stop candidate is the architecture immediately before
the first removal whose measured loss exceeds $\tau = \sqrt{p(1-p)/m}$;
$\tau$ is a heuristic based on the estimated standard error of a
single accuracy, and it does not bound the cumulative loss relative to
the full model. The remaining $B - 1$ shortlist places
go to the candidates with the highest masked accuracies at sizes not
yet taken. Each shortlisted candidate starts from a separate copy of
the initial model with excluded components removed and continues
training under the dataset's standard settings, with epochs and early
stopping counted from the continuation start; the five initial epochs
are charged additionally in every reported cost. The full architecture
is a candidate whenever the elimination path marks it; the empty
architecture is never a candidate and serves as the baseline
$a(\emptyset)$. The final choice is the trained candidate with the
highest validation accuracy, ties toward the stop candidate. Only this
final comparison uses fully trained models; all intermediate removals
use the masked game.

\begin{algorithm}[t]
\caption{\method-neighbor: architecture selection}
\label{alg:neighbor-selection}
\begin{algorithmic}[1]
\Require Full model with neighborhoods $\mathcal{N}$;
         training and validation sets $D_{\mathrm{train}},D_{\mathrm{val}}$;
         candidate budget $1 \leq B \leq |\mathcal{N}|$
\Ensure Selected trained model

\State Train the full model for five epochs, obtaining weights $\theta_0$
\State Define $a(S)$ as its validation accuracy with neighborhoods
       outside $S$ masked and weights fixed at $\theta_0$
\State Cache all evaluations of $a(S)$
\State $m \gets |D_{\mathrm{val}}|$,
       $p \gets a(\mathcal{N})$,
       $\tau \gets \sqrt{p(1-p)/m}$

\State $S \gets \mathcal{N}$, $\mathcal{K} \gets \{\mathcal{N}\}$,
       $S_{\mathrm{stop}} \gets \varnothing$

\For{$k=|\mathcal{N}|,\ldots,2$}
    \State $i^* \gets \arg\max_{i \in S} a(S \setminus \{i\})$
    \State $S' \gets S \setminus \{i^*\}$
    \If{$S_{\mathrm{stop}}=\varnothing$
        \textbf{and} $a(S)-a(S')>\tau$}
        \State $S_{\mathrm{stop}} \gets S$
               \Comment{Mark the candidate before this removal}
    \EndIf
    \State $\mathcal{K} \gets \mathcal{K} \cup \{S'\}$;\quad $S \gets S'$
           \Comment{Continue even after marking a candidate}
\EndFor

\If{$S_{\mathrm{stop}}=\varnothing$}
    \State $S_{\mathrm{stop}} \gets S$
           \Comment{Use the singleton if no loss exceeded $\tau$}
\EndIf

\State $\mathcal{T} \gets \{S_{\mathrm{stop}}\} \cup \{B-1$ candidates
       in $\mathcal{K}\setminus\{S_{\mathrm{stop}}\}$ with highest
       $a(S)\}$

\For{each $T' \in \mathcal{T}$}
    \State Initialize $f_{T'}$ from a separate copy of $\theta_0$,
           removing components outside $T'$
    \State Continue training $f_{T'}$ to completion under standard
           settings
    \State $q(T') \gets \operatorname{Accuracy}(f_{T'},D_{\mathrm{val}})$
\EndFor

\State \Return $f_{T^*}$, $T^* = \arg\max_{T' \in \mathcal{T}} q(T')$
       (ties toward $S_{\mathrm{stop}}$)
\end{algorithmic}
\end{algorithm}

\section{Datasets and Training Settings}
\label{app:setup}

\subsection{Datasets, Splits, and Liftings}
\label{app:datasets}
\label{app:coverage}

Our experiments use three groups of datasets
(\cref{tab:dataset-stats}). Benzene, Fluoride-Carbonyl,
Alkane-Carbonyl, and Mutagenicity are binary graph-classification
tasks from the GraphXAI suite \citep{agarwal2023graphxai}, built on
the attribution benchmarks of \citet{sanchezlengeling2020evaluating};
each molecule's label is determined by a known functional group, so
every graph carries a ground-truth explanation. MUTAG, NCI1, NCI109,
and PROTEINS are graph-classification benchmarks from the TUDataset
collection \citep{morris2020tudataset}; Cora, Citeseer, and Pubmed
\citep{yang2016planetoid} are citation networks with node-level
classification (one complex each; transductive, random 50/25/25 node
split, split seed 0). MANTRA \citep{ballester2025mantra} is a
collection of triangulated 2-manifolds, natively simplicial; we use
the orientability task. The ZINC subset
\citep{dwivedi2023benchmarking} (regression, scored by mean absolute
error) appears only in the exhaustive six-neighborhood benchmark of
\cref{app:benchmark}, under two training configurations described in
\cref{app:training}; both use ZINC's predefined subset splits of
10{,}000/1{,}000/1{,}000 graphs (train/validation/test, about 23 nodes
and 50 edges per graph). It is not part of the selection or
explanation experiments.

\begin{table}[t]
\centering
\small
\setlength{\tabcolsep}{4pt}
\caption{\textbf{Dataset statistics} after preprocessing. Cells per rank are means over complexes (rank 0 nodes, rank 1 edges, rank 2 2-cells; MANTRA is natively simplicial). Node-level datasets are one complex whose splits partition the nodes (transductive, random 50/25/25 split, split seed 0); their split columns count nodes.}
\label{tab:dataset-stats}
\begin{tabular}{lrrrrrrr}
\toprule
Dataset & Train & Val & Test & Cells r0 & r1 & r2 & Classes \\
\midrule
\multicolumn{8}{l}{\emph{GraphXAI benchmarks (lifted to cellular complexes)}} \\
Benzene & 8400 & 1800 & 1800 & 21 & 22 & 2 & 2 \\
Fluoride-Carbonyl & 6069 & 1300 & 1302 & 21 & 23 & 2 & 2 \\
Alkane-Carbonyl & 787 & 168 & 170 & 21 & 23 & 2 & 2 \\
Mutagenicity & 1237 & 265 & 266 & 29 & 30 & 2 & 2 \\
\midrule
\multicolumn{8}{l}{\emph{Graph benchmarks (lifted to cellular complexes)}} \\
MUTAG & 94 & 47 & 47 & 18 & 20 & 3 & 2 \\
NCI1 & 2055 & 1027 & 1028 & 30 & 32 & 4 & 2 \\
NCI109 & 2063 & 1031 & 1033 & 30 & 32 & 4 & 2 \\
PROTEINS & 556 & 278 & 279 & 39 & 73 & 33 & 2 \\
Cora & 1354 & 677 & 677 & 2,708 & 5,278 & 1,648 & 7 \\
Citeseer & 1663 & 831 & 833 & 3,327 & 4,552 & 1,259 & 6 \\
Pubmed & 9858 & 4929 & 4930 & 19,717 & 44,324 & 9,325 & 3 \\
\midrule
\multicolumn{8}{l}{\emph{Natively higher-order (simplicial)}} \\
MANTRA (orientability) & 21568 & 10785 & 10785 & 10 & 34 & 23 & 2 \\
\bottomrule
\end{tabular}
\end{table}

\textbf{Liftings.} Graph datasets are lifted to cellular complexes
with TopoBench's cycle lifting \citep{telyatnikov2024topobench}:
chordless cycles of up to ten nodes become rank-2 cells. Two feature
liftings appear. The \emph{standard} lifting (ProjectionSum) gives
each higher-rank cell the sum of its constituent nodes' features. The
\emph{structure-only} lifting gives every cell above rank 0 a constant
all-ones feature vector, identical across cells, at the width of the
rank below (MANTRA's native convention); rank-1 and rank-2 cells are
treated the same way, node features are kept, and the complex itself
is unchanged. MANTRA needs no lifting.

\textbf{Participation.} The prediction-explanation experiments of
\cref{sec:predictions} use the four GraphXAI datasets (GNN models on
graphs; GCCN models on their lifted complexes), the four unannotated
TU datasets for faithfulness, and MANTRA for the model diagnostics.
The architecture-selection experiments of \cref{sec:performance} use
eleven datasets (all of \cref{tab:dataset-stats} except PROTEINS,
which appears in the faithfulness study only). On Pubmed, HOPSE
preprocessing does not complete at practical cost (its eigensolvers
exceed eight GPU-hours on the lifted complex), so Pubmed carries no
HOPSE runs; the HOPSE sweep covers the complex-level datasets.

\subsection{Models, Training, and Explainer Settings}
\label{app:training}

\textbf{TNN models (architecture selection).} All GCCN selection
experiments use the \texttt{topotune\_nonlinear\_exact} configuration:
a TopoTune GCCN backbone with two layers, hidden width 32, one
two-layer GCN sub-network per neighborhood (BatchNorm, ReLU), and a
linear sum-pool readout. HOPSE models replace each neighborhood's
message-passing module with fixed encodings of that neighborhood's
graph plus a learned MLP block (\cref{def:hopse}): four encodings per
neighborhood (Laplacian-eigenvector positional, random-walk
structural, electrostatic positional, and heat-kernel-diagonal
structural), each targeting 20 dimensions per cell and concatenated
with the cell's input feature; the encoder width is 64, with two
backbone layers and a sum-pool readout. Both families train with Adam
(learning rate $10^{-3}$, no weight decay), the learning rate halved
every 50 epochs, up to 300 epochs with early stopping (patience 25) on
the dataset's monitored validation metric, and batch size 64 for graph
datasets, 256 for MANTRA, and full-batch for the citation
complexes. Checkpoint selection is validation-best. Continuations in
the selection pipeline restart these settings from the five-epoch
initial model.

\textbf{ZINC benchmark configurations.} The exhaustive benchmark's two
ZINC settings share ZINC's predefined subset splits and are scored by
negative test MAE. The fixed-recipe setting uses the shared GCCN
configuration above (26{,}112 backbone parameters at the full
six-neighborhood model); the strong setting replaces the GCN
sub-networks with GIN sub-networks (104{,}448 backbone parameters).

\textbf{GNN models (graph explanation).} Parameter-matched GCN
\citep{kipf2017semisupervised} and GIN \citep{xu2018how} models: three
layers, input width 14, hidden width 32, two classes; Adam with
learning rate $10^{-3}$ and weight decay $5 \times 10^{-4}$;
class-weighted cross-entropy; batch size 64; up to 300 epochs with
early stopping on validation balanced accuracy (patience 30);
checkpoint = best validation balanced accuracy; stratified shuffle
splits with seed 42; three training seeds per model.

\textbf{Baseline explainers} (settings from the saved suite): SubgraphX
with 10 Monte-Carlo tree-search rollouts, minimum subgraph size 3,
$c_{\mathrm{puct}} = 10$, 14 expansion atoms, local radius 4, and 100
reward samples per evaluation; GNNExplainer with learning rate 0.05
for 1000 epochs (selected on training data; the GraphXAI defaults of
0.01/200 were also run); PGExplainer trained once per dataset-model
pair on up to 1000 training graphs (30 epochs, learning rate 0.003
selected on training data; defaults 10 epochs/0.1 also run). Every
method receives the annotated motif's size and is evaluated on the
same instances. Occlusion scores each cell by the leave-one-out drop
$v(P) - v(P \setminus \{i\})$ under the same masked game;
gradient-times-input scores each cell by the gradient of the target
margin with respect to its raw input features, dotted with those
features and summed over the feature dimension (signed); the random
baseline draws one seeded uniform score per cell. All methods are
ranked by signed score, descending.

\textbf{Seeds, hardware, software.} Comparisons share seeds 42--44;
the GCCN $B{=}2$ selection additionally carries seeds 45--46
(\cref{tab:performance-summary}). All local runs use one NVIDIA H200
GPU per job; PyTorch 2.3.0 (CUDA 12.1) and a pinned TopoBench commit,
listed in \cref{app:repro}.

\section{Additional Results for \method-cell}
\label{app:cell-results}

\subsection{Graph Explanation Accuracy and Runtime}
\label{app:graph-results}
\label{app:explanations}
\label{app:tables}

\textbf{From scores to node sets.} On graphs, every method is scored
as a node set of the annotated motif's size $b$: the attribution
output takes the $b$ highest-scoring nodes, the explanation output
takes the greedy candidate $\hat{S}_{b}$, and baselines are converted
with the same top-$b$ rule (GNNExplainer and PGExplainer score a node
by the sum of its incident edges' mask weights; SubgraphX scores a
node by its best explored-coalition reward, plus one for membership in
the returned subgraph; ties break toward the lower node index). On
lifted complexes, sizes are matched at the cell level instead: the
size-$b$ cell set is projected to the union of its constituent nodes
without re-truncation, so the projected node set can exceed $b$ nodes
and the Jaccard score penalizes the excess. GEA is the best Jaccard
over the instance's admissible annotated candidates. The evaluated
population is every ground-truth-positive complex in the test split,
shared by every method: 917 (Benzene), 231 (Fluoride-Carbonyl), 96
(Mutagenicity), and 58 (Alkane-Carbonyl) instances.

\begin{table}[t]
\centering
\small
\caption{\textbf{Explanation accuracy by model type.} Node-level GEA (mean $\pm$ sd, 3 seeds) for each parameter-matched model type, one panel per type. The zero-fill row runs our explanation search with feature zero-filling in place of detachment. Bold: column best; gray: within one sd.}
\label{tab:by-subject}
\setlength{\tabcolsep}{4pt}
\begin{tabular}{lcccc}
\toprule
\emph{GCN models} & Benzene & Fluoride & Mutagenicity & Alkane \\
\midrule
\method{} (attribution) & 0.467 $\pm$ 0.078 & \cellcolor{black!10}0.398 $\pm$ 0.040 & \cellcolor{black!10}\textbf{0.723 $\pm$ 0.074} & \cellcolor{black!10}\textbf{0.411 $\pm$ 0.058} \\
\method{} (explanation) & 0.377 $\pm$ 0.012 & 0.289 $\pm$ 0.094 & 0.317 $\pm$ 0.062 & 0.318 $\pm$ 0.074 \\
\method{} (explanation, zero-fill) & \cellcolor{black!10}0.648 $\pm$ 0.055 & 0.244 $\pm$ 0.034 & 0.561 $\pm$ 0.062 & \cellcolor{black!10}0.378 $\pm$ 0.075 \\
SubgraphX & \cellcolor{black!10}\textbf{0.664 $\pm$ 0.027} & 0.226 $\pm$ 0.024 & 0.344 $\pm$ 0.036 & 0.276 $\pm$ 0.013 \\
PGExplainer & \cellcolor{black!10}0.642 $\pm$ 0.044 & \cellcolor{black!10}\textbf{0.429 $\pm$ 0.092} & 0.290 $\pm$ 0.102 & 0.203 $\pm$ 0.137 \\
GNNExplainer & 0.443 $\pm$ 0.054 & 0.210 $\pm$ 0.014 & 0.152 $\pm$ 0.033 & 0.066 $\pm$ 0.050 \\
Random & 0.250 $\pm$ 0.000 & 0.219 $\pm$ 0.000 & 0.132 $\pm$ 0.000 & 0.173 $\pm$ 0.000 \\
\bottomrule
\end{tabular}
\par\smallskip
\begin{tabular}{lcccc}
\toprule
\emph{GIN models} & Benzene & Fluoride & Mutagenicity & Alkane \\
\midrule
\method{} (attribution) & 0.859 $\pm$ 0.009 & \cellcolor{black!10}0.507 $\pm$ 0.047 & \cellcolor{black!10}\textbf{0.798 $\pm$ 0.124} & \cellcolor{black!10}\textbf{0.500 $\pm$ 0.081} \\
\method{} (explanation) & \cellcolor{black!10}\textbf{0.969 $\pm$ 0.009} & 0.340 $\pm$ 0.010 & 0.480 $\pm$ 0.135 & 0.354 $\pm$ 0.125 \\
\method{} (explanation, zero-fill) & 0.569 $\pm$ 0.064 & 0.313 $\pm$ 0.038 & 0.273 $\pm$ 0.164 & 0.391 $\pm$ 0.028 \\
SubgraphX & 0.796 $\pm$ 0.029 & 0.439 $\pm$ 0.107 & 0.274 $\pm$ 0.055 & 0.296 $\pm$ 0.010 \\
PGExplainer & 0.453 $\pm$ 0.239 & \cellcolor{black!10}\textbf{0.555 $\pm$ 0.067} & 0.175 $\pm$ 0.200 & 0.205 $\pm$ 0.160 \\
GNNExplainer & 0.256 $\pm$ 0.081 & 0.166 $\pm$ 0.035 & 0.158 $\pm$ 0.023 & 0.162 $\pm$ 0.049 \\
Random & 0.250 $\pm$ 0.000 & 0.219 $\pm$ 0.000 & 0.132 $\pm$ 0.000 & 0.173 $\pm$ 0.000 \\
\bottomrule
\end{tabular}

\end{table}

\textbf{Runtime protocol.} \Cref{tab:runtime} reports measured
wall-clock seconds per explained instance, averaged over both model
types and three seeds, on one NVIDIA H200 GPU per run, with
each method timed end to end on the same instances; PGExplainer's
one-time per-dataset training is reported as its own row.

\begin{table}[t]
\centering
\small
\caption{\textbf{Per-instance explanation cost.} Measured wall-clock seconds per explained instance (mean $\pm$ sd over both model types and 3 seeds), on one NVIDIA H200 per run. $^{\dagger}$PGExplainer's per-instance entry is inference only; the last row is its one-time training.}
\label{tab:runtime}
\begin{tabular}{lcccc}
\toprule
Method & Benzene & Fluoride & Mutagenicity & Alkane \\
\midrule
\method{} (attribution) & 3.3 $\pm$ 0.9 & 3.3 $\pm$ 0.8 & 4.0 $\pm$ 3.1 & 3.3 $\pm$ 0.8 \\
\method{} (explanation) & 0.33 $\pm$ 0.11 & 0.35 $\pm$ 0.11 & 0.67 $\pm$ 1.05 & 0.36 $\pm$ 0.11 \\
SubgraphX & 21.1 $\pm$ 16.1 & 34.3 $\pm$ 28.9 & 51.6 $\pm$ 79.4 & 40.4 $\pm$ 28.8 \\
PGExplainer$^{\dagger}$ & 0.002 $\pm$ 0.000 & 0.002 $\pm$ 0.000 & 0.002 $\pm$ 0.000 & 0.002 $\pm$ 0.000 \\
GNNExplainer & 0.81 $\pm$ 0.08 & 0.80 $\pm$ 0.08 & 0.80 $\pm$ 0.07 & 0.82 $\pm$ 0.08 \\
Random & 0.001 $\pm$ 0.001 & 0.003 $\pm$ 0.010 & 0.001 $\pm$ 0.001 & 0.002 $\pm$ 0.003 \\
PGExplainer training (one-time)$^{\dagger}$ & 1500.8 $\pm$ 159.5 & 1081.3 $\pm$ 115.9 & 221.3 $\pm$ 21.7 & 142.5 $\pm$ 15.5 \\
\bottomrule
\end{tabular}
\end{table}

\subsection{Feature Lifting and Motif Recovery}
\label{app:lifting-diagnostics}

\Cref{tab:lifting-diagnostics} reports the lifting diagnostics behind
\cref{sec:predictions-lifted}, on the explanation output
$\hat{S}_{b}$ of the GCCN with all eleven neighborhoods. Face retention is the
fraction of instances whose selected coalition contains at least one
rank-2 cell; boundary recall is the mean recall of the annotated
motif's rank-1 cells. The two statistics are computed separately, and
both use the full ground-truth-positive test populations and three
training seeds. The final column re-scores the
standard lifting with its summed higher-rank features recomputed on
every masked complex, so a retained face carries only the features of
its retained nodes; everything else (weights, players, target) is
unchanged. The difference in motif recovery between the two liftings
persists when the summed features are recomputed after masking.

\begin{table}[t]
\centering
\small
\setlength{\tabcolsep}{4pt}
\caption{\textbf{Lifting diagnostics for the explanation output.} Face retention: fraction of instances whose selected coalition contains at least one rank-2 cell. Boundary recall: mean recall of the annotated motif's rank-1 cells. Both are computed on the full ground-truth-positive test populations (mean $\pm$ sd over three training seeds). The last column rescores the standard lifting with its summed features recomputed on every masked complex instead of held fixed (node-projected GEA); its structural-lifting counterpart is in \cref{tab:lifted}.}
\label{tab:lifting-diagnostics}
\resizebox{\linewidth}{!}{%
\begin{tabular}{lccccc}
\toprule
 & \multicolumn{2}{c}{Face retention} & \multicolumn{2}{c}{Boundary recall} & GEA, recomputed feat. \\
\cmidrule(lr){2-3}\cmidrule(lr){4-5}
Dataset & standard & structure-only & standard & structure-only & (standard lifting) \\
\midrule
Benzene & 0.84 $\pm$ 0.14 & 0.95 $\pm$ 0.05 & 0.40 $\pm$ 0.14 & 0.86 $\pm$ 0.06 & 0.56 $\pm$ 0.09 \\
Fluoride-Carbonyl & 0.00 $\pm$ 0.00 & 0.00 $\pm$ 0.00 & 0.16 $\pm$ 0.06 & 0.53 $\pm$ 0.04 & 0.28 $\pm$ 0.02 \\
Mutagenicity & 0.05 $\pm$ 0.04 & 0.13 $\pm$ 0.15 & 0.08 $\pm$ 0.04 & 0.30 $\pm$ 0.04 & 0.18 $\pm$ 0.02 \\
Alkane-Carbonyl & 0.02 $\pm$ 0.03 & 0.03 $\pm$ 0.05 & 0.07 $\pm$ 0.13 & 0.50 $\pm$ 0.24 & 0.25 $\pm$ 0.17 \\
\bottomrule
\end{tabular}}
\end{table}

\subsection{Faithfulness Without Annotations}
\label{app:faithfulness}

Deletion-AUC detaches cells cumulatively, most important first. With
$m_{k}$ the target margin after detaching the $k$ highest-ranked of
$n$ cells and $s = \operatorname{sign}(m_{0})$ orienting the margin
toward the predicted class,
\[
\mathrm{AUC}
= \sum_{k=1}^{n} \frac{\sigma(s\,m_{k-1}) + \sigma(s\,m_{k})}{2}
  \cdot \frac{1}{n},
\]
the trapezoid rule applied to the predicted-class probability
($\sigma$: logistic function) over the removed fraction, from the full
complex to full removal. Lower is better. The explained models are
GCCNs with all eleven neighborhoods under the structure-only lifting;
explanations target the model's own prediction. The evaluated
population is a fixed seed-0 draw of 50 test complexes per dataset,
except MUTAG, which uses its entire 47-complex test split.

\begin{table}[t]
\centering
\small
\caption{\textbf{Faithfulness without ground truth.} Deletion-AUC (lower is better; a fixed draw of 50 test complexes per dataset, except MUTAG, which uses its whole 47-complex test split). Subjects are GCCNs with all eleven neighborhoods under the structure-only lifting; explanations target the model's own prediction.}
\label{tab:nogt}
\begin{tabular}{lcccc}
\toprule
 & NCI1 & NCI109 & MUTAG & PROTEINS \\
\midrule
\method{} & \cellcolor{black!10}\textbf{0.218 $\pm$ 0.005} & \cellcolor{black!10}\textbf{0.203 $\pm$ 0.044} & \cellcolor{black!10}\textbf{0.175 $\pm$ 0.043} & \cellcolor{black!10}\textbf{0.338 $\pm$ 0.067} \\
Occlusion & 0.254 $\pm$ 0.021 & \cellcolor{black!10}0.240 $\pm$ 0.038 & 0.231 $\pm$ 0.006 & 0.431 $\pm$ 0.065 \\
Gradient$\times$Input & 0.552 $\pm$ 0.016 & 0.507 $\pm$ 0.049 & 0.499 $\pm$ 0.073 & 0.638 $\pm$ 0.047 \\
Random & 0.545 $\pm$ 0.004 & 0.491 $\pm$ 0.035 & 0.481 $\pm$ 0.014 & 0.666 $\pm$ 0.058 \\
\bottomrule
\end{tabular}
\end{table}

\subsection{MANTRA: Attribution Patterns and Feature Ablations}
\label{app:mantra}

\textbf{Metrics.} The overall AUROC in \cref{sec:predictions-nogt} is
computed on MANTRA's full test split from each model's
orientable-class score. The count-conditional AUROC restricts
comparisons to pairs of test complexes with identical $(V, E, F)$ cell
counts: over all count groups containing both classes (four groups),
it is the pooled probability that an orientable complex outscores a
non-orientable one within the same group; between-group pairs are
excluded, so a model that predicts from counts alone cannot exceed
0.5. These AUROCs are model diagnostics and are distinct from the
balanced test accuracy used for selection results.

\textbf{Feature ablations.} \Cref{tab:mantra-features} collects the
comparison behind \cref{sec:predictions-nogt}: models restricted to
cell counts or purely structural features, the GCCN and HOPSE
selections of \cref{sec:performance-selection}, and a GCCN trained
with Hodge Laplacian eigenvectors concatenated to its input features.
The rows differ in model-selection procedure and seed count (stated
per row in the table), so the table compares feature sets across
configurations rather than isolating one controlled feature change.

\begin{table}[t]
\centering
\small
\caption{\textbf{MANTRA orientability: feature and model comparison.} Balanced test accuracy. The count-only row is a single logistic regression fit; the structural-features and eigenvector rows are means over three training seeds whose per-seed values were not exported, so no sd is shown. The feature study trains its own GCCN configuration, distinct from the selected models, so rows compare feature sets across configurations. Selection rows report mean $\pm$ sd over seeds.}
\label{tab:mantra-features}
\begin{tabular}{lcc}
\toprule
Model / input features & Balanced test acc. & Runs \\
\midrule
Logistic regression on cell counts & 0.505 & 1 fit \\
Structural features only & 0.541 & 3 seeds (means only) \\
GCCN, selected ($B{=}2$, plain-accuracy selection) & 0.641 $\pm$ 0.123 & 5 seeds \\
GCCN + Hodge Laplacian eigenvectors & 0.876 & 3 seeds (means only) \\
HOPSE, selected ($B{=}2$) & 0.771 $\pm$ 0.017 & 3 seeds \\
HOPSE, sweep best & 0.792 $\pm$ 0.009 & 3 seeds \\
\bottomrule
\end{tabular}
\end{table}

\section{Additional Results for \method-neighbor}
\label{app:neighbor-results}

\subsection{Architecture Selection, Baselines, and Compute}
\label{app:selection}

\begin{table}[t]
\centering
\small
\setlength{\tabcolsep}{3pt}
\caption{\textbf{Architecture selection: full results by model family.} Test accuracy (mean $\pm$ sd; GCCN $B{=}2$ uses five seeds 42--46, all other columns the shared seeds 42--44; $^{b}$balanced test accuracy on MANTRA) and, in parentheses, total training FLOPs as a multiple of that dataset's $B{=}1$ run (\cref{app:flops-conventions}); sweeps are charged their full grid. Gap = selection $B{=}2$ minus that family's sweep; sweeps use per-seed validation-best configurations. ---: not run. Pubmed carries no HOPSE runs (\cref{app:datasets}).}
\label{tab:performance-summary}
\begin{tabular}{lcccc}
\toprule
\multicolumn{5}{c}{GCCN family} \\[1pt]
Dataset & $B{=}1$ & $B{=}2$ & sweep & gap \\
\midrule
Benzene & 0.925 $\pm$ 0.007 {\scriptsize($\times$1.0)} & 0.926 $\pm$ 0.004 {\scriptsize($\times$2.0)} & 0.926 $\pm$ 0.002 {\scriptsize($\times$22.7)} & +0.000 \\
Fluoride-Carb. & 0.901 $\pm$ 0.057 {\scriptsize($\times$1.0)} & 0.939 $\pm$ 0.006 {\scriptsize($\times$1.9)} & 0.938 $\pm$ 0.004 {\scriptsize($\times$20.6)} & +0.001 \\
Mutagenicity & 0.967 $\pm$ 0.011 {\scriptsize($\times$1.0)} & 0.975 $\pm$ 0.010 {\scriptsize($\times$2.0)} & 0.964 $\pm$ 0.014 {\scriptsize($\times$15.4)} & +0.012 \\
Alkane-Carb. & 0.965 $\pm$ 0.006 {\scriptsize($\times$1.0)} & 0.962 $\pm$ 0.005 {\scriptsize($\times$2.4)} & 0.975 $\pm$ 0.003 {\scriptsize($\times$16.2)} & -0.012 \\
NCI1 & 0.764 $\pm$ 0.011 {\scriptsize($\times$1.0)} & 0.758 $\pm$ 0.010 {\scriptsize($\times$2.8)} & 0.759 $\pm$ 0.022 {\scriptsize($\times$15.0)} & -0.000 \\
MUTAG & 0.879 $\pm$ 0.025 {\scriptsize($\times$1.0)} & 0.906 $\pm$ 0.024 {\scriptsize($\times$2.1)} & 0.901 $\pm$ 0.025 {\scriptsize($\times$13.1)} & +0.006 \\
NCI109 & 0.758 $\pm$ 0.011 {\scriptsize($\times$1.0)} & 0.756 $\pm$ 0.009 {\scriptsize($\times$1.8)} & 0.747 $\pm$ 0.015 {\scriptsize($\times$11.3)} & +0.009 \\
Cora & 0.838 $\pm$ 0.008 {\scriptsize($\times$1.0)} & 0.835 $\pm$ 0.005 {\scriptsize($\times$2.3)} & 0.856 $\pm$ 0.006 {\scriptsize($\times$14.9)} & -0.021 \\
Citeseer & 0.673 $\pm$ 0.014 {\scriptsize($\times$1.0)} & 0.670 $\pm$ 0.007 {\scriptsize($\times$1.8)} & 0.683 $\pm$ 0.002 {\scriptsize($\times$15.1)} & -0.013 \\
Pubmed & 0.860 $\pm$ 0.004 {\scriptsize($\times$1.0)} & 0.863 $\pm$ 0.004 {\scriptsize($\times$2.3)} & 0.865 $\pm$ 0.001 {\scriptsize($\times$14.4)} & -0.002 \\
MANTRA$^{b}$ & 0.532 $\pm$ 0.028 {\scriptsize($\times$1.0)} & 0.641 $\pm$ 0.123 {\scriptsize($\times$1.8)} & 0.548 $\pm$ 0.000 {\scriptsize($\times$11.0)} & +0.093 \\
\bottomrule
\end{tabular}
\par\medskip
\begin{tabular}{lccc}
\toprule
\multicolumn{4}{c}{HOPSE family} \\[1pt]
Dataset & $B{=}2$ & sweep & gap \\
\midrule
Benzene & 0.934 $\pm$ 0.001 {\scriptsize($\times$5.4)} & 0.936 $\pm$ 0.003 {\scriptsize($\times$36.4)} & -0.002 \\
Fluoride-Carb. & 0.918 $\pm$ 0.004 {\scriptsize($\times$8.5)} & 0.934 $\pm$ 0.009 {\scriptsize($\times$41.6)} & -0.015 \\
Mutagenicity & 0.961 $\pm$ 0.008 {\scriptsize($\times$5.9)} & 0.976 $\pm$ 0.009 {\scriptsize($\times$36.3)} & -0.015 \\
Alkane-Carb. & 0.969 $\pm$ 0.009 {\scriptsize($\times$10.5)} & 0.963 $\pm$ 0.007 {\scriptsize($\times$47.8)} & +0.006 \\
NCI1 & 0.759 $\pm$ 0.024 {\scriptsize($\times$5.6)} & 0.745 $\pm$ 0.011 {\scriptsize($\times$34.5)} & +0.013 \\
MUTAG & 0.844 $\pm$ 0.012 {\scriptsize($\times$10.9)} & 0.844 $\pm$ 0.054 {\scriptsize($\times$53.3)} & +0.000 \\
NCI109 & 0.720 $\pm$ 0.005 {\scriptsize($\times$5.2)} & 0.737 $\pm$ 0.011 {\scriptsize($\times$27.1)} & -0.017 \\
Cora & 0.851 $\pm$ 0.005 {\scriptsize($\times$10.2)} & --- & --- \\
Citeseer & 0.712 $\pm$ 0.007 {\scriptsize($\times$6.8)} & --- & --- \\
MANTRA$^{b}$ & 0.771 $\pm$ 0.017 {\scriptsize($\times$9.6)} & 0.792 $\pm$ 0.009 {\scriptsize($\times$68.7)} & -0.021 \\
\bottomrule
\end{tabular}
\end{table}

\textbf{Sweeps.} The GCCN sweep is a TopoTune-style neighborhood grid
of 22--25 configurations per dataset: the saved record's per-dataset
draws on the eight datasets it covered, and one fixed 23-configuration
draw reused verbatim on the three datasets swept for this paper
(per-dataset counts in \cref{fig:lead}'s legend; the configuration
lists ship with the released pipeline). The HOPSE sweep is its
published 12-configuration grid. Sweep winners are selected per seed
by validation accuracy and charged the full grid.

\begin{table}[t]
\centering
\small
\caption{\textbf{Shortlist baselines with shared initialization and training settings.} Test accuracy (mean $\pm$ sd, shared seeds 42--44; $^{b}$balanced on MANTRA). All three columns share the five-epoch initial model and standard continuation; random $B{=}2$ replaces backward elimination with two uniformly random architectures, and the full-model default trains all 11 neighborhoods with no search.}
\label{tab:shortlist-baselines}
\begin{tabular}{lccc}
\toprule
Dataset & \method-neighbor $B{=}2$ & Random $B{=}2$ & Full default \\
\midrule
Benzene & 0.927 $\pm$ 0.000 & 0.925 $\pm$ 0.013 & 0.927 $\pm$ 0.003 \\
Fluoride-Carbonyl & 0.939 $\pm$ 0.007 & 0.929 $\pm$ 0.002 & 0.938 $\pm$ 0.003 \\
Mutagenicity & 0.976 $\pm$ 0.012 & 0.932 $\pm$ 0.061 & 0.972 $\pm$ 0.015 \\
Alkane-Carbonyl & 0.965 $\pm$ 0.006 & 0.973 $\pm$ 0.003 & 0.955 $\pm$ 0.007 \\
NCI1 & 0.757 $\pm$ 0.013 & 0.755 $\pm$ 0.026 & 0.766 $\pm$ 0.005 \\
MUTAG & 0.915 $\pm$ 0.021 & 0.879 $\pm$ 0.025 & 0.879 $\pm$ 0.012 \\
NCI109 & 0.755 $\pm$ 0.011 & 0.755 $\pm$ 0.011 & 0.749 $\pm$ 0.001 \\
Cora & 0.839 $\pm$ 0.002 & 0.831 $\pm$ 0.021 & 0.836 $\pm$ 0.003 \\
Citeseer & 0.669 $\pm$ 0.001 & 0.660 $\pm$ 0.013 & 0.666 $\pm$ 0.010 \\
Pubmed & 0.862 $\pm$ 0.004 & 0.853 $\pm$ 0.006 & 0.861 $\pm$ 0.002 \\
MANTRA$^{b}$ & 0.615 $\pm$ 0.103 & 0.687 $\pm$ 0.114 & 0.777 $\pm$ 0.035 \\
\bottomrule
\end{tabular}
\end{table}

\textbf{Shortlist controls.} Each control starts from the same
five-epoch initial model and continues training under the standard
settings of the selection pipeline. The random $B{=}2$ control draws
two architectures uniformly at random from the 2047 non-empty
neighborhood subsets (draw seeded by dataset and seed, redrawing
duplicates), continues both, and selects between them by validation
accuracy: two continuations, like the selection pipeline, but without
backward elimination's 66 validation passes. The full-model default
continues one model, the all-neighborhood architecture, with no search
and no selection step.

\textbf{Compute accounting.}
\label{app:flops-conventions}
All FLOPs are counted under one convention: PyTorch's
\texttt{FlopCounterMode} with dense pricing, explicit formulas for
operations the counter does not price (eigendecomposition, SVD, sparse
products), and every stage charged, including HOPSE's encoding
preprocessing. Runs that predate the profiler are priced by rebuilding
the recorded architecture, profiling one training epoch and the
evaluation passes under the same convention, and combining these
per-epoch counts with the recorded epoch and evaluation tallies; we
describe such costs as profiler-based reconstructions. Exact
neighborhood attribution requires all $2^{11} = 2048$ masked
validation passes; backward elimination requires $(n^{2}+n)/2 = 66$;
sampled cell attribution requires at most $Tn + 1$ forward passes per
instance, with $T$ counting every evaluated ordering
(\cref{app:cell-intervention,tab:runtime}). Wall-clock comparisons run
on one NVIDIA H200 GPU per run.

\begin{figure}[t]
  \centering
  \includegraphics[width=\linewidth]{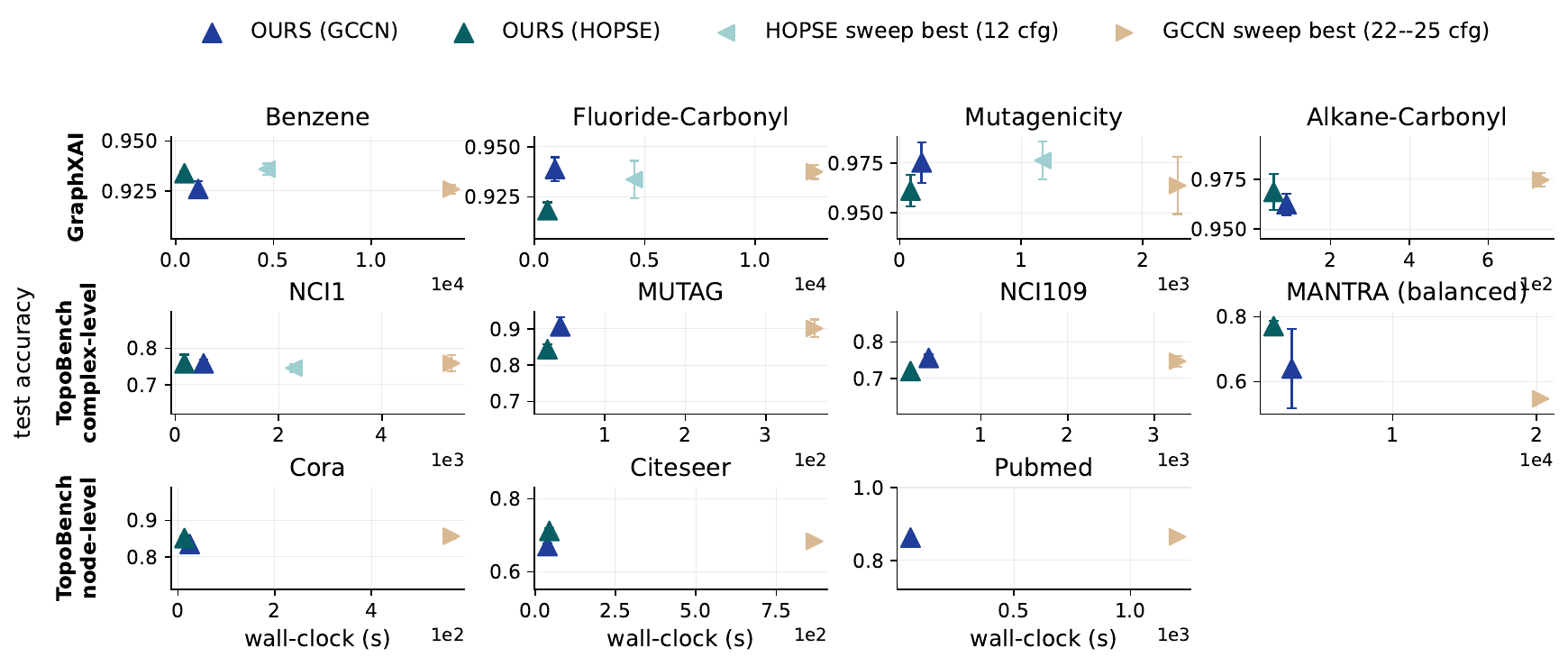}
  \caption{\textbf{Wall-clock variant of \cref{fig:lead}.} The same
    four marker groups as \cref{fig:lead} (selected GCCN, selected
    HOPSE, and each family's sweep), with the horizontal axis showing
    measured wall-clock seconds (one NVIDIA H200 GPU per run) instead
    of FLOPs. Wall-clock time includes HOPSE's preprocessing.}
  \label{fig:app-lead-wall}
\end{figure}

\subsection{Sensitivity Analyses}
\label{app:sensitivity}

\textbf{Initial-training length.} \Cref{tab:stem-sensitivity} repeats
the $B{=}2$ selection with the initial model trained for 3, 5
(default), or 10 epochs. Mean selected test accuracy moves by at most
0.012 on Benzene and NCI1 and by 0.036 on MUTAG; the selected
neighborhood sets themselves differ across settings.

\begin{table}[t]
\centering
\small
\caption{\textbf{Sensitivity to the brief-run epoch count.} Test accuracy of the $B{=}2$ selection (mean $\pm$ sd, 3 seeds) with the initial model trained for 3, 5 (default), or 10 epochs.}
\label{tab:stem-sensitivity}
\begin{tabular}{lccc}
\toprule
Brief-run epochs & Benzene & NCI1 & MUTAG \\
\midrule
3 & 0.926 $\pm$ 0.008 & 0.763 $\pm$ 0.006 & 0.879 $\pm$ 0.033 \\
5 (protocol) & 0.927 $\pm$ 0.000 & 0.757 $\pm$ 0.013 & 0.915 $\pm$ 0.021 \\
10 & 0.925 $\pm$ 0.006 & 0.769 $\pm$ 0.008 & 0.887 $\pm$ 0.033 \\
\bottomrule
\end{tabular}
\end{table}

\textbf{Selection metric on MANTRA.} A separately run three-seed
variant replaces ordinary validation accuracy with balanced validation
accuracy in the masked game, the threshold's $p$, and the final
comparison ($\tau$'s binomial form is then an approximation, since
balanced accuracy is not a binomial proportion). It selects 11, 11,
and 9 neighborhoods across seeds 42--44 and reaches $0.804 \pm 0.011$
balanced test accuracy, against $0.615 \pm 0.103$ for the original
plain-accuracy run on the same seeds, consistent with ordinary
validation accuracy rewarding majority-class predictions on this
imbalanced dataset. \Cref{fig:lead,tab:performance-summary} report the
original plain-accuracy selection.

\begin{figure}[p]
  \centering
  \includegraphics[width=0.96\linewidth]{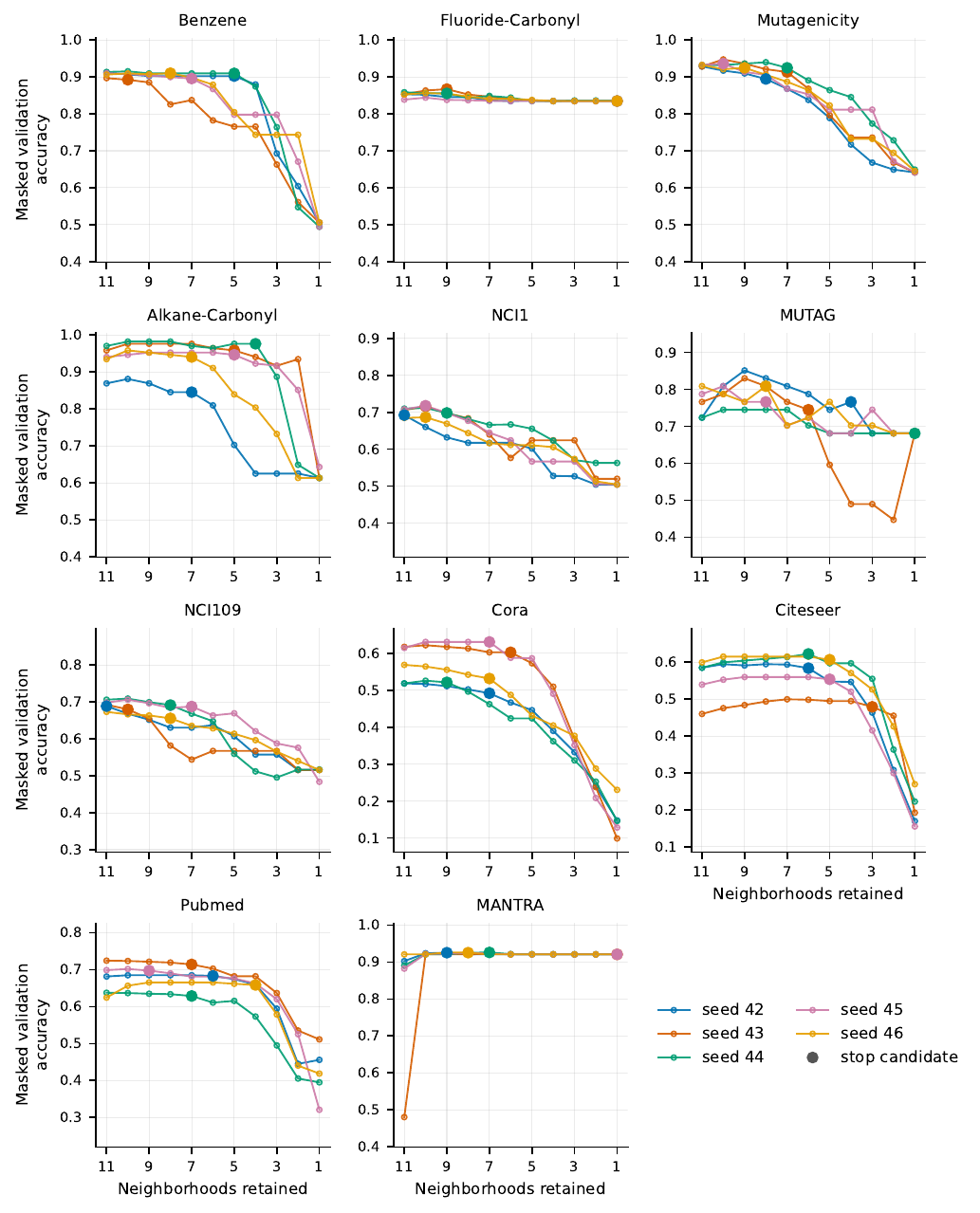}
  \caption{\textbf{Backward-elimination traces on all eleven
    datasets.} Masked validation accuracy along the GCCN elimination
    path, from eleven neighborhoods to one, for training seeds
    42--46. Markers identify the stop candidate selected before the
    first removal whose accuracy loss exceeds $\tau$; elimination
    continues after this candidate is marked. All panels share one
    $y$-axis span, positioned per panel. MANTRA uses ordinary
    validation accuracy in this experiment.}
  \label{fig:elimination-traces}
\end{figure}

\subsection{Exhaustive Benchmark and Attribution Agreement}
\label{app:benchmark}

The controlled evaluation of \cref{sec:performance-controlled} uses
all $2^{6} = 64$ subsets of the first six neighborhoods of \cref{app:selection-pipeline}, each trained separately from scratch to the standard budget, under twelve dataset/hyperparameter settings and
seeds 42--44; the empty subset is a trained model with node features only. The twelve settings are the four GraphXAI molecule datasets, MUTAG, NCI1, NCI109, the two ZINC configurations of
\cref{app:training}, and the three citation graphs. Seven settings were available while the selection procedure was developed; five (MUTAG, NCI109, Alkane-Carbonyl, and the two ZINC configurations) were held out and first evaluated after the procedure was fixed.

\label{app:attr-agreement}
\Cref{tab:attr-agreement} compares exact Shapley attributions over the
six neighborhoods computed two ways: from masked evaluations of the
five-epoch model (the logit column scores validation logits, the
primary comparison; the accuracy column re-scores the masked side by
validation accuracy) and from the separately trained subsets' test
values. Agreement is the Spearman correlation between seed-averaged
attribution profiles; stability is the mean $\pm$ sd of the per-seed
correlations, which mixes seed-to-seed variation of the masked profile
into the same statistic and is reported separately for that reason.
The pooled row correlates $12 \times 6 = 72$ points (one per setting
and neighborhood, each a three-seed mean); intervals are 95\%
bootstrap over the twelve setting blocks (10{,}000 resamples), and the
primary statistic's permutation $p$ is $0.048$ (20{,}000
permutations). The two games differ in split and score (validation
logits or accuracy versus test accuracy or negative MAE), a mismatch
the saved record does not allow us to remove.

\begin{table}[t]
\centering
\small
\caption{\textbf{Attribution agreement with retraining.} Spearman correlation between masked-game and retraining-game Shapley profiles over the six neighborhoods (masked: logit-scored, validation split; retraining: test metric). Agreement: correlation between seed-averaged attribution profiles. Stability: per-seed correlation, mean $\pm$ sd (seeds 42--44). Pooled row: 95\% CI in brackets.}
\label{tab:attr-agreement}
\begin{tabular}{lccc}
\toprule
Dataset & Agreement $\rho$ (logit) & (accuracy) & Per-seed $\rho$ (stability) \\
\midrule
MUTAG & -0.37 & +0.89 & -0.16 $\pm$ 0.74 \\
NCI1 & -0.31 & +0.14 & -0.12 $\pm$ 0.95 \\
NCI109 & +0.09 & +0.09 & +0.20 $\pm$ 0.70 \\
ZINC-subset (fixed recipe) & -0.09 & -0.09 & +0.14 $\pm$ 0.29 \\
ZINC-subset (strong) & -0.31 & -0.31 & -0.45 $\pm$ 0.14 \\
Alkane-Carbonyl & +0.49 & +0.49 & +0.50 $\pm$ 0.36 \\
Benzene & +0.83 & +0.31 & +0.43 $\pm$ 0.37 \\
Citeseer & +0.66 & +0.66 & +0.56 $\pm$ 0.12 \\
Cora & +0.94 & +0.94 & +0.94 $\pm$ 0.00 \\
PubMed & +0.54 & +0.37 & +0.39 $\pm$ 0.22 \\
Fluoride-Carbonyl & -0.09 & +0.77 & +0.26 $\pm$ 0.45 \\
Mutagenicity & +0.94 & +0.77 & +0.71 $\pm$ 0.20 \\
\midrule
Pooled (72 points) & +0.23 [-0.05, +0.52] & +0.32 [+0.06, +0.60] & \\
\bottomrule
\end{tabular}
\end{table}

\subsection{Oracle Evaluation of Candidate Generation}
\label{app:selection-evidence}

\Cref{tab:oracle-generation} reruns backward elimination with each
removal scored by the benchmark's seed-averaged test values, per
setting. Because it consults test scores, this analysis is an oracle
diagnostic of the candidate generator; it is not a validation-only
application of the practical method, and it does not validate the
masked scores the practical pipeline uses. On the five held-out
settings the greedy path contains the benchmark's best architecture in
four, with worst regret $+0.0068$ (the GIN-backbone ZINC recipe); on
the seven development settings it matches the best architecture's
test value in five (on NCI1 with a size-5 subset rather than the
size-3 optimum) and trails by at most $+0.0042$.

\begin{table}[t]
\centering
\small
\setlength{\tabcolsep}{4pt}
\caption{\textbf{Oracle evaluation of candidate generation on the exhaustive benchmark.} Backward elimination is rerun with each removal scored by the separately trained coalitions' seed-averaged test values (three seeds); ties break toward the first neighborhood in the order of \cref{app:selection-pipeline}. Because it consults test scores, this is a diagnostic of the candidate generator, not a deployable selection. Empty = trained model with node features only; Full = all six neighborhoods; Best = best of all 64 subsets, with its size; Path best = best subset on the greedy elimination path; Regret = Best $-$ Path best. ZINC values are negative test MAE (higher is better).}
\label{tab:oracle-generation}
\begin{tabular}{llcccccc}
\toprule
Setting & Status & Empty & Full & Best ($k$) & Path best ($k$) & Regret \\
\midrule
Benzene & development & 0.739 & 0.921 & 0.925 (3) & 0.925 (3) & +0.0000 \\
Mutagenicity & development & 0.821 & 0.956 & 0.957 (5) & 0.957 (5) & +0.0000 \\
Alkane-Carbonyl & held-out & 0.967 & 0.967 & 0.984 (4) & 0.984 (4) & +0.0000 \\
Fluoride-Carbonyl & development & 0.822 & 0.935 & 0.943 (4) & 0.943 (4) & +0.0000 \\
MUTAG & held-out & 0.723 & 0.901 & 0.922 (5) & 0.922 (5) & +0.0000 \\
NCI1 & development & 0.667 & 0.759 & 0.768 (3) & 0.768 (5) & +0.0000 \\
NCI109 & held-out & 0.680 & 0.756 & 0.762 (3) & 0.762 (3) & +0.0000 \\
ZINC-subset (fixed recipe)$^{m}$ & held-out & -0.699 & -0.548 & -0.534 (5) & -0.534 (5) & +0.0000 \\
ZINC-subset (strong)$^{m}$ & held-out & -0.723 & -0.225 & -0.177 (4) & -0.184 (5) & +0.0068 \\
Cora & development & 0.632 & 0.848 & 0.861 (3) & 0.861 (3) & +0.0000 \\
Citeseer & development & 0.644 & 0.681 & 0.688 (3) & 0.685 (3) & +0.0024 \\
PubMed & development & 0.854 & 0.861 & 0.867 (3) & 0.863 (5) & +0.0042 \\
\bottomrule
\end{tabular}
\par\smallskip\footnotesize $^{m}$negative test MAE.
\end{table}

\section{Reproduction Instructions}
\label{app:repro}
\label{app:reproducibility}

\textbf{Environment.} Python 3.11; PyTorch 2.3.0 (CUDA 12.1);
\texttt{torch-geometric} with matching scatter/sparse wheels; TopoBench
pinned at the commit recorded in the repository. The repository's
README gives the exact installation commands.

\textbf{Released code.} The release contains the explainer library with
its test suite, including the exactness check of the Shapley
implementation against brute-force enumeration, a tutorial, and the
experiment runners and configuration files used in this paper.

\textbf{Rerunning experiments.} Each experiment has one runner under
\texttt{experiments/} (for example, architecture selection, the sweeps,
the shortlist controls, the balanced-selection MANTRA variant, and the
faithfulness study), with seeds and output locations set by
command-line arguments. The exhaustive six-neighborhood benchmark and
the GraphXAI explainer suite were produced on a separate cluster; their
configurations and saved numerical results are included with the
release.

\end{document}